%% file: usenix.tex
\documentclass[letterpaper,twocolumn,10pt]{article}

\usepackage[font=footnotesize,labelfont=bf]{caption}
\usepackage{usenix,epsfig,endnotes}
\usepackage{xcolor}
\usepackage{microtype}
\usepackage{graphicx}
\usepackage{booktabs} 
\usepackage{subcaption}
\usepackage{xspace}
\usepackage{balance}
\usepackage{hyperref}
\usepackage{fancyhdr}
\usepackage{extramarks}
\usepackage{amsmath}
\usepackage{amsthm}
\usepackage{amsfonts}
\usepackage{tikz}
\usepackage{amssymb}
\usepackage{adjustbox}

\usepackage{algorithm} 
\usepackage[noend]{algpseudocode}
\algnewcommand\algorithmicforeach{\textbf{for each}}

\algdef{S}[FOR]{ForEach}[1]{\algorithmicforeach\ #1\ \algorithmicdo}

\newcommand{\claire}[1]{\textcolor{purple}{\{Claire: #1\}}}

\usepackage{varwidth} 
\usepackage{semantic}
\usepackage{pifont}
\newcommand{\cmark}{\ding{51}}%
\newcommand{\xmark}{\ding{55}}%
\newcommand{\name}{Nazar\xspace}

\newcommand{\eg}{e.g.,\xspace}
\newcommand{\ie}{i.e.,\xspace}

\newcommand{\OOD}{data drift\xspace}

\usepackage{enumitem}
\setitemize{noitemsep,topsep=0pt,parsep=0pt,partopsep=0pt}
\setenumerate{noitemsep,topsep=0pt,parsep=0pt,partopsep=0pt}
\setdescription{itemsep=0pt,topsep=0pt,parsep=0pt,partopsep=0pt,itemindent=0em,leftmargin=0.5em}

\newenvironment{denseitemize}{
\begin{itemize}[topsep=2pt, partopsep=0pt, leftmargin=1.5em]
  \setlength{\itemsep}{2pt}
  \setlength{\parskip}{0pt}
  \setlength{\parsep}{0pt}
}{\end{itemize}}

\newenvironment{denseenum}{
\begin{enumerate}[topsep=2pt, partopsep=0pt, leftmargin=1.5em]
  \setlength{\itemsep}{2pt}
  \setlength{\parskip}{0pt}
  \setlength{\parsep}{0pt}
}{\end{enumerate}}

\usepackage{titling}
\begin{document}
\date{}
\title{\Large \bf\name: Monitoring and Adapting ML Models on Mobile Devices}
\author{
{\rm Wei Hao}\\
Columbia University
\and
{\rm Zixi Wang}\\
Columbia University
\and
{\rm Lauren Hong}\\
Columbia University
\and
{\rm Lingxiao Li}\\
Columbia University
\and
{\rm Nader Karayanni}\\
Columbia University
\and
{\rm Chengzhi Mao}\\
Columbia University
\and
{\rm Junfeng Yang}\\
Columbia University
\and
{\rm Asaf Cidon}\\
Columbia University
}
\maketitle
\thispagestyle{empty}
\input{abstract}
\input{intro}
\input{background}
\input{overview}
\input{detection}

\input{driftanalysis}
\input{adaptation}
\input{implementation}

\input{evaluation} 
\input{conclusions}

{\footnotesize \bibliographystyle{acm}
\bibliography{usenix}}


\end{document}

%% file: abstract.tex
\begin{abstract}

ML models are increasingly being pushed to mobile devices, for low-latency inference and offline operation. However, once the models are deployed, it is hard for ML operators to track their accuracy, which can degrade unpredictably (\eg due to data drift). We design \name, the first end-to-end system for  continuously monitoring and adapting models on mobile devices without requiring feedback from users. Our key observation is that often model degradation is due to a specific root cause, which may affect a large group of devices. Therefore, once \name detects a consistent degradation across a large number of devices, it employs a root cause analysis to determine the origin of the problem, and  applies a cause-specific adaptation.
We evaluate \name on two computer vision datasets, and show 
it consistently boosts accuracy compared to existing approaches. On a dataset containing photos collected from driving cars, \name improves the accuracy on average by 15\%. 
\end{abstract}

%% file: intro.tex
\section{Introduction} 

Companies such as Google, Meta, Apple and LinkedIn, are increasingly running their ML models \emph{on users' mobile devices}, so they can run inference locally at a low latency, even when the device is offline. Examples include text suggestion~\cite{ramaswamy2019federated,chen2019federated}, ranking posts, object detection and tracking for virtual reality~\cite{wu2019machine}, speech recognition~\cite{paulik2021federated} and targeted advertising~\cite{wang2023flint}.
As state-of-the-art ML models are very large, they are typically trained in the companies' datacenters with fleets of GPUs~\cite{wu2019machine,qiu2022ml}. A team of \emph{ML operators} is usually responsible for retraining the models and pushing new model versions to the users' mobile devices.

However, in large-scale settings, ML operator teams lack sufficient visibility into the performance of the ML models running on users devices~\cite{shankar2022operationalizing}. 
Model accuracy across devices is highly variable over time, and can change due to unexpected shifts in the input data (termed \emph{data drift}) or hardware issues in specific devices (\eg low-quality cameras, microphones)~\cite{cidon2021characterizing,qiu2022ml}. 
Although the ML operators may fine-tune or retrain models to improve accuracy, it is quite difficult today to detect \OOD in the first place or verify whether accuracy has been restored properly after fine-tuning~\cite{shankar2022operationalizing,qiu2022ml,DBLP:journals/corr/abs-2108-13557,shankar2022rethinking}. The crux is that after the models are deployed, the ML operators have no ``ground truth'' on how accurate their models are because users do not manually label data.

Although many algorithms from the ML literature can detect data drift \cite{MSP, energy-ood, DBLP:journals/corr/NguyenYC14,DBLP:journals/corr/abs-1812-05720, OE, odin, MD, ssl-ood, CSI, lee2017training,rabanser2019failing,alibi-detect}, and adapt to the drifts \cite{
li2017learning,rebuffi2017icarl,shmelkov2017incremental,yin2018feature,maltoni2019continuous}, they often operate with impractical assumptions (\eg assume fully-labeled data, or a single source of data drift, or require significant computational resources or manual inspection), and have never been tested in a real end-to-end system.

We present \name, the first end-to-end system for continuously monitoring and addressing \OOD in large-scale ML deployments on mobile devices. We explicitly design \name to operate fully automatically without human involvement so that it can practically support large-scale ML deployments.  It consists of three main functions: (a) automatically \emph{detect} when \OOD occurs on mobile devices, (b) \emph{analyze} the root cause of the \OOD, and (c) \emph{adapt} to the cause that triggered the \OOD, all without requiring any user input. Given that user-labeled data is not available and that operators have little resources for manual diagnosis, \name employs \emph{self-supervised} methods for detecting, analyzing, and adapting to \OOD, which we elaborate further below.



\textbf{Detection.} After considering a wide range of detection algorithms, we decided to adopt a \emph{confidence threshold} method that detects \OOD by comparing the model's confidence of the predictions against a threshold. This method requires no user-labeled data or manually supplied surrogate models, and is thus suitable for \name's automation requirement while providing good accuracy. Moreover, it is computationally lightweight and can easily work locally on any mobile devices.

\textbf{Root cause analysis.} A classical approach~\cite{diff,apriori} for root cause analysis is called \emph{frequent itemset mining}, which finds sets of attributes (in our case, root causes) that are tied with specific data types (detected \OOD). While this approach is useful in uncovering causes of drift, it leads to many duplicate root causes, which are subsets of or overlap with each other, and therefore is often used in settings in which a human manually inspects the results and acts upon them~\cite{diff}. It is untenable in large-scale settings that \name targets. Hence, we introduce novel \emph{set reduction} and \emph{counterfactual analysis} algorithms, two heuristics that automatically produce a small set of likely \OOD root causes by eliminating root causes that subsume each other and overlap.

\textbf{By-cause adaptation.} The vast majority of existing adaptation techniques assume labeled data, and even those that do not, make completely impractical assumptions. They assume a single source of data drift, and adapt the model on each newly-arriving input~\cite{DBLP:journals/corr/abs-2006-10726,DBLP:journals/corr/abs-2110-09506}, which leads to poor accuracy when there are multiple sources of drift. Instead, we propose a novel \emph{by-cause} model adaptation method, which selectively adapts the model only to the specific cause of accuracy degradation with the data that is suspected to have it.

The three \name components work harmoniously together, continuously reinforcing each other in an end-to-end loop. As our experiments show, \name's analysis accurately identifies the root cause of the drift, so that by-cause adaption is applied only to data drifted for the same cause. Otherwise, a model adapted to one cause is unlikely to have high accuracy on data that drifted due to completely different causes. Once an adapted model is deployed, \name's detection will leverage the new model's confidence to detect drifts, continuously improving accuracy. These components together enable \name to effectively obtain a consistently and significantly higher accuracy than existing approaches that adapt on all inputs, 
without requiring any manual work.


We implement \name on Amazon AWS, and evaluate it on two streaming datasets: cityscapes, a self-driving car dataset composed of images taken from driving vehicles in various European cities~\cite{cityscapes}, and an ImageNet-derived dataset that emulates a species classification app. 
\name provides an average of 15\% and up to 22\% of accuracy boost on all data and an average of 19\% and up to 50\% on drifted data compared to the baselines on cityscapes. 
We will open source our datasets and code upon publication.

Our main technical contributions are:
\begin{denseenum}
\item First end-to-end online system for model monitoring and adaptation for mobile devices.
\item Fully-automated root cause analysis mechanism that introduces set reduction and counterfactual analysis to effectively narrow down the root causes of model degradation.
\item New self-supervised adaptation technique that adapts by root cause, which provides a significant accuracy boost over existing approaches when deployed with \name's effective root cause analysis, without any human input.
\end{denseenum}

%% file: background.tex
\section{Background and Related Work}
\label{sec:background}

Companies are increasingly deploying models on user devices for faster inference and to support low connectivity settings. On-device models support a wide range of use cases, including object recognition~\cite{wu2019machine}, text auto-complete~\cite{ramaswamy2019federated, chen2019federated} and ranking posts and ads~\cite{wu2019machine,wang2023flint}.
\vspace{-.5em}
\paragraph{Where to train the models?}
These models can be trained either centrally in the cloud, or in a distributed fashion across the devices themselves (termed \emph{federated learning}). 
The advantages of training the model in the cloud are: (a) the model training and evaluation process is simpler, and (b) powerful datacenter-based AI compute capabilities (\eg a large number of GPUs/TPUs) are available for training. An advantage of federated learning is that it does not require uploading training data to the cloud, facilitating better privacy.
While federated learning has seen increased interest both in academia~\cite{oort,fedscale,ghosh,fedml} and in industry~\cite{wang2023flint,ramaswamy2019federated, chen2019federated}, in this work, we primarily focus on the more common scenario, where the models are trained in the cloud. \name's principles and ideas can also be applied to the federated learning case. 

\vspace{-.5em}
\paragraph{Model compression and quantization.} Conducting inference on devices introduces several significant challenges. 
Since mobile devices are quite heterogeneous, and may impose significant resource limitations, models need to be adapted to be run on those devices. Several systems attempt to address this challenge, including TensorFlow Lite~\cite{tflite}, an extension to TensorFlow that adapts trained models to resource-limited devices (\eg using quantization). Another example is Mistify~\cite{mistify}, which automatically adapts the model's architecture to fit the device's constraints.
\vspace{-.5em}
\paragraph{Accuracy degradation.} However, techniques like model compression can introduce subtle changes in the models, which may degrade their accuracy. These discrepancies in accuracy across devices can impact the applications' quality of service~\cite{cidon2021characterizing,qiu2022ml}, and lead to security vulnerabilities~\cite{hao2022tale, qiu2022ml}. For example, while quantization can reduce the model size exponentially to fit on resource-constrained devices, 
it can also lead to worse accuracy for specific classes~\cite{cidon2021characterizing}. This accuracy degradation is hard to anticipate in advance. 
\vspace{-.5em}
\paragraph{Data drift.} Data drift is a classic problem that affects models that operate on streaming data, where the distribution of the newly-arriving data diverges from training data's distribution~\cite{widmer1996learning,mccloskey1989catastrophic,french1999catastrophic}. 
Data drift can occur for many reasons: temporal environmental changes or corruptions affecting the model input (\eg weather, noise), changes in the input distribution itself (\eg a self-driving model trained in the US might not recognize street signs in India), and sensor-related issues (\eg problems with device microphone or camera lense).
To deal with this problem, models are often frequently retrained with fresh data, even if some older data is retained, so their training data better reflects the types of inputs they will encounter during inference~\cite{ekya,parisi2019continual}.
The data drift problem is exacerbated in on-device deployments, because: (a) different devices may see divergent data distributions, and (b) when retraining in large-scale deployments, it is not feasible to retrain on data from \emph{all devices}. Therefore, one needs to decide when to retrain and which devices to sample from.  

Ekya~\cite{ekya} is a recent system that tackles a variant of this problem, by jointly scheduling retraining and inference on the same ``edge'' servers. In the case of Ekya, an edge server is a server with multiple GPUs sitting at the network’s edge (\eg in a content distribution network), unlike our setting where inference is done on resource-constrained mobile devices. A second key difference is that, unlike \name, Ekya assumes labels are available for adapting the models.
\vspace{-.5em}
\paragraph{Lack of visibility and monitoring.}
There are several efforts to provide more visibility and monitoring in ML pipelines. ML-EXray~\cite{qiu2022ml} is a framework that helps ``debug'' models by allowing ML operators to instrument logs and debugging assertions. It targets pre-processing issues, quantization bugs, and suboptimal kernels. Another data validation system for ML pipelines implemented at Meta~\cite{shankar2023moving} tries to identify possible attributes that caused model performance drop, from the training input. In contrast to these systems, \name focuses on post-deployment issues caused by data drift, while these systems focus more on the pre-deployment validation setting. 

In summary, while there are existing systems that identify problems in models pre-deployment, and existing ad-hoc detection and adaptation techniques, to the best of our knowledge there is no automated end-to-end system for monitoring and adapting models on mobile devices.


%% file: overview.tex
\section{Design} 

\subsection{Overview}

\begin{figure}[t!]
\centering
\includegraphics[width=\columnwidth]{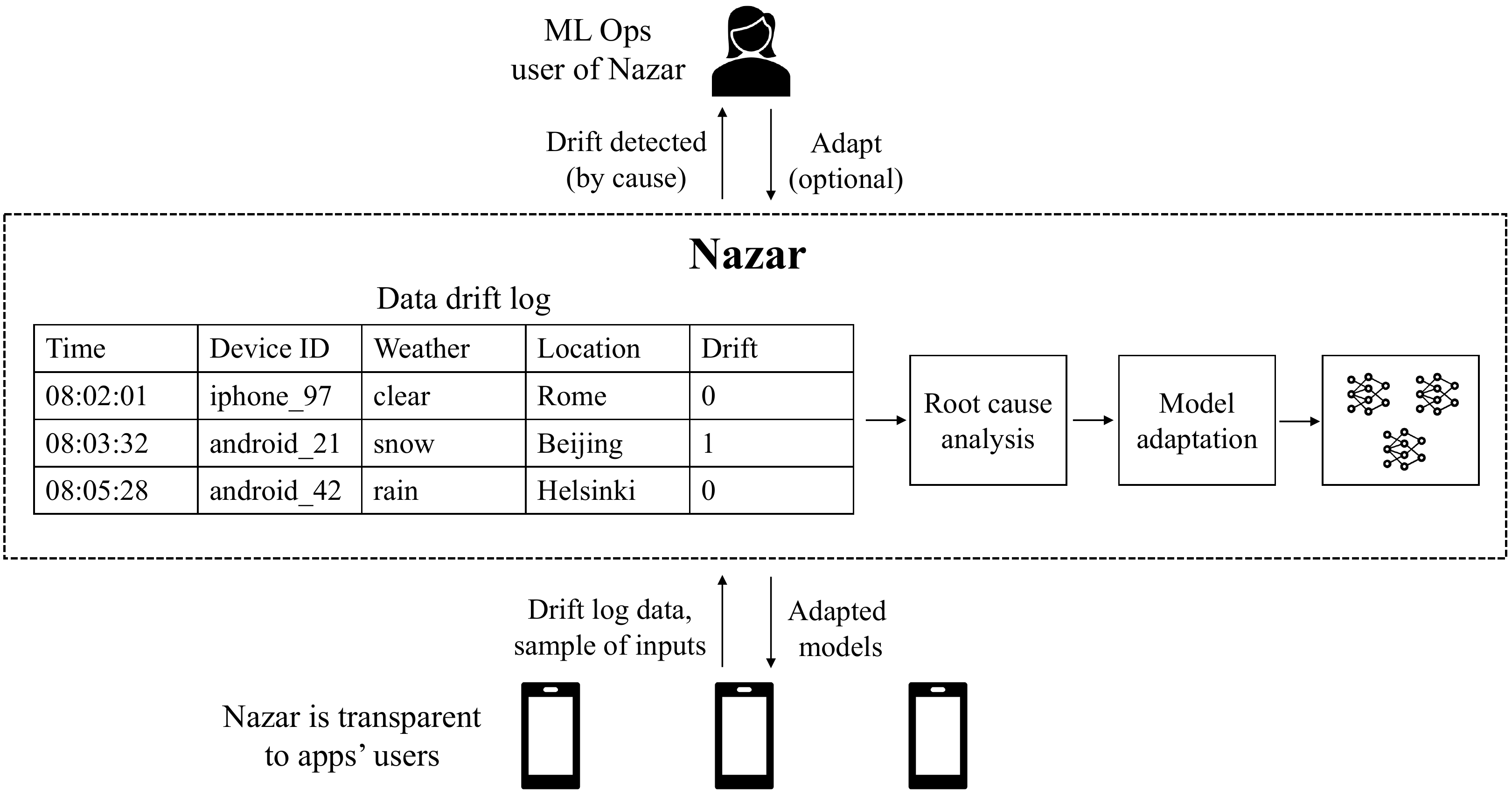}
\vspace{-1em}
\caption{\name's design.}
\vspace{-1em}
\label{fig:design}
\end{figure}

We present an overview of \name's design (Figure~\ref{fig:design}). 
We design \name to be an end-to-end system that can automatically monitor data drift, identify the main root causes for the drift, and adapt models to them at scale, while dynamically evolving with changing data distributions in the real world.
\vspace{-.5em}
\paragraph{Usage modes.} The user of \name is the ML ops team, which maintains and retrains models. ML ops can choose their level of interaction with \name: by default, \name runs in an ``autopilot'' mode, where monitoring, analysis and adaptation are all done automatically. The ML ops team can choose to more closely interact with the system, receiving alerts when data drift occurs, and manually deciding when to trigger the adaptation and on which root causes to apply them.
\vspace{-.5em}
\paragraph{On-device detection.} App users (\ie users of the apps that run the AI models) are completely oblivious to \name. \name operates in the cloud, and interacts with the on-device models using an API. The apps contribute two types of information to \name. First, each time the models conduct inference, they also subsequently run a very lightweight drift detection algorithm (described in \S\ref{sec:detect}). The result of this algorithm, along with additional metadata about the device and model (\eg the device's ID, current location, local time, model version) are sent to \name in the cloud. We call this metadata collected from the devices a \emph{drift log entry}. Second, periodically, in the background, the device will sample some percentage (by default 1\%) of the inputs to the model, and send them to the cloud for model adaptation purposes.
\vspace{-.5em}
\paragraph{Root cause analysis.} The drift log entries are ingested into a database in the cloud, called the \emph{drift log}. Periodically, \name runs a \emph{root cause analysis} algorithm (\S\ref{sec:analysis}). The algorithm is composed of: (a) a series of database queries on the data drift log to identify potential sets of causes of data drift, and (b) \emph{set reduction} and \emph{counterfactual analysis} algorithms, which reduce redundant causes and rank them to select the most important ones to adapt to. When drift is indeed detected, \name generates an alert for the ML ops team.
\vspace{-.5em}
\paragraph{By-cause adaptation.} After drift is detected, \name generates new \emph{by-cause model adaptations} that adapt to the specific root causes of the drift (\S\ref{sec:adaptation}). These adaptations are pushed to the app users' devices and used for new inferences. Since model updates can accumulate, \name consolidates the different versions to capture the relevant root causes with a small number of model adaptations. 
For inference, the device chooses which model version to use for each input, by selecting the one with attributes that best match the input.
\vspace{-.5em}
\paragraph{Evolving drift detection.}
In the process of detecting, analyzing and adapting to drift, the concept of what is non-drift and what is drift evolves over time. If \name adapts on a certain source of drift, it should not continue to be detected as drift over and over again. 
%
%
Therefore, ideally the detector and root cause analysis should keep ``adapting'' to the new distribution of data, and only exhibit a high detection rate if they see data that diverges from the data the original model was trained on and that \name adapted its models to.  
%
%

\vspace{-.5em}
\paragraph{Design principles.}
The main questions in designing \name is how to design its detection, root cause analysis and adaptation mechanisms. Two primary principles guide our design:
\begin{denseenum}
\item As \name is transparent to app users, it requires \emph{self-supervised} methods that cannot rely on labeled data.
\item \name must support fully-automated operation, without relying on the ML ops team instruction. 
\end{denseenum}

%% file: detection.tex
\begin{table*}[!t]
\centering
\footnotesize
\begin{tabular}{|l|r|r|r|r|r|r|r|r|r|r|}
\hline

 & \multicolumn{1}{|l|}{Threshold~\cite{MSP}} & \multicolumn{1}{|l|}{KS-test~\cite{rabanser2019failing}} & \multicolumn{1}{|l|}{OE~\cite{OE}} & \multicolumn{1}{|l|}{Odin~\cite{odin}} & \multicolumn{1}{|l|}{MD~\cite{MD}}& \multicolumn{1}{|l|}{SSL~\cite{ssl-ood}}&
\multicolumn{1}{|l|}{CSI~\cite{CSI}}&
\multicolumn{1}{|l|}{GOdin~\cite{Godin}}\\
\hline
No secondary dataset & \cmark & \cmark & \xmark & \xmark & \xmark & \cmark & \cmark &  \cmark \\
\hline 
No secondary model & \cmark & \cmark & \cmark & \cmark & \cmark & \xmark & \xmark &  \cmark\\
\hline
No backpropagation & \cmark & \cmark & \cmark & \xmark & \cmark & \cmark & \cmark &  \xmark\\
\hline
No batching & \cmark & \xmark & \cmark & \cmark & \cmark & \cmark & \cmark  & \cmark\\
\hline
\end{tabular}
\caption{Comparison of different data drift detection algorithms from the ML literature. \name uses the threshold method, which simply applies a threshold on the model's softmax output, such as the max softmax score.}
\vspace{-1.5em}
\label{tab:detection-comparison}
\end{table*}

\subsection{On-Device Data Drift Detection} \label{sec:detect}
We now describe \name's data drift detector, which runs on the mobile devices. The goal of the data drift detector is to detect when data drift has occurred on a particular device in a lightweight fashion, without requiring any user input.

\subsubsection{Data Drift Detection Techniques}

When designing \name's drift detection mechanism, we considered different \OOD (also known as ``concept drift'' and ``out-of-distribution drift'') algorithms from the ML literature. 
However, most of these techniques were not designed for an on-device inference setting. 
Table~\ref{tab:detection-comparison} summarizes the eight primary techniques we evaluated.
We introduce the classes of techniques, and describe our rationale for focusing on methods that are derived from the model's inference output.
\vspace{-.5em}
\paragraph{Detecting based on model output.}
One class of \OOD detection techniques assumes that \OOD is correlated with the model being ``unsure'' about the correct class to output. Therefore, they apply various metrics on the model's \emph{logit} vector (or un-normalized log probabilities). As a refresher, in deep neural network (DNN) models, the model outputs an array that assigns a probability for each class. Typically the highest scoring class is chosen as the predicted class.
The distribution of this vector can indicate how ``certain'' a model is about its prediction.
For example, the MSP \emph{threshold}~\cite{MSP} checks if the maximum softmax value of the logit vector is below a certain threshold. If it is, the model is uncertain about the final prediction, which is correlated with \OOD. Prior work also proposes other types of thresholds or metrics, such as the entropy of the softmax values~\cite{alibi-detect} and other similar scores~\cite{energy-ood}. 
In our experiments, we found these thresholds to perform almost identically to MSP. MSP has the small advantage that it is normalized between 0 and 1, making it a simpler knob to tune, and hence we adopt it. 
The advantage of this class of algorithms is that they can simply be applied to the model's inference output with negligible computational overhead (since the logit scores are anyways computed by the inference), and do not require any outside information. 

\vspace{-.5em}
\paragraph{Statistical test on batch of outputs.}  Orthogonal to the aforementioned methods that use a simple threshold for \OOD detection, statistical tests can be combined with any of these scores to determine whether the inference data diverges from the training data distribution~\cite{alibi-detect}. 
For example, Kolmogorov-Smirnov (KS) test (KS-test in Table~\ref{tab:detection-comparison})
compares the empirical cumulative distribution functions (CDFs) of the two datasets and calculates their maximum difference.
Prior work~\cite{rabanser2019failing} conducts empirical studies on using different combinations of statistic test and finds that applying KS-test on softmax outputs yields the best detection accuracy. Hence, this is the main statistical test we evaluate.

\vspace{-.5em}
\paragraph{External datasets.}
Another class of methods assumes training-time access to a dataset that contains drifted data samples. The intuition is that such a ``drift dataset'' will help train a model to be more sensitive to all drifts. Examples include Outlier-Exposer~\cite{OE} (OE in Table~\ref{tab:detection-comparison}), Odin~\cite{odin} and Mahalanobis Distance~\cite{MD}  (MD). We deem these approaches impractical for our setting, because our data is unlabeled, and we cannot assume that users will prepare drift datasets which is a meaningless concept for a user.

\vspace{-.5em}
\paragraph{Secondary model.}
Some techniques employ auxiliary models to detect \OOD. SSL~\cite{ssl-ood} and CSI~\cite{CSI} co-train an auxiliary model for self-supervised tasks on a transformed version of the dataset (\eg they rotate images in the original dataset and perform rotation angle classification). The assumption is that if there was no drift, after training, both models would have high confidence on the same input. 
We rule out this class of methods, given that some mobile devices are resource constrained, and we cannot rely on them to invoke an additional model purely for \OOD detection. 

\vspace{-.5em}
\paragraph{Methods that require backpropagation.}
Generalized Odin~\cite{Godin} (GOdin) is an expanded version of Odin that does not require the use of a secondary dataset 
but requires adding small adversarial perturbations to make the model more confident when data is in distribution, and less confident when data has drifted. However, to add these perturbations, it requires an extra step of \emph{backpropagation} after the softmax values are read, \ie the neural network needs to be traversed in reverse. Unfortunately backpropagation cannot be used in certain compressed models, which means this technique would not be usable for many types of adapted models pushed to user devices. 

\begin{figure}[t]
\centering
\includegraphics[width=0.8\columnwidth]{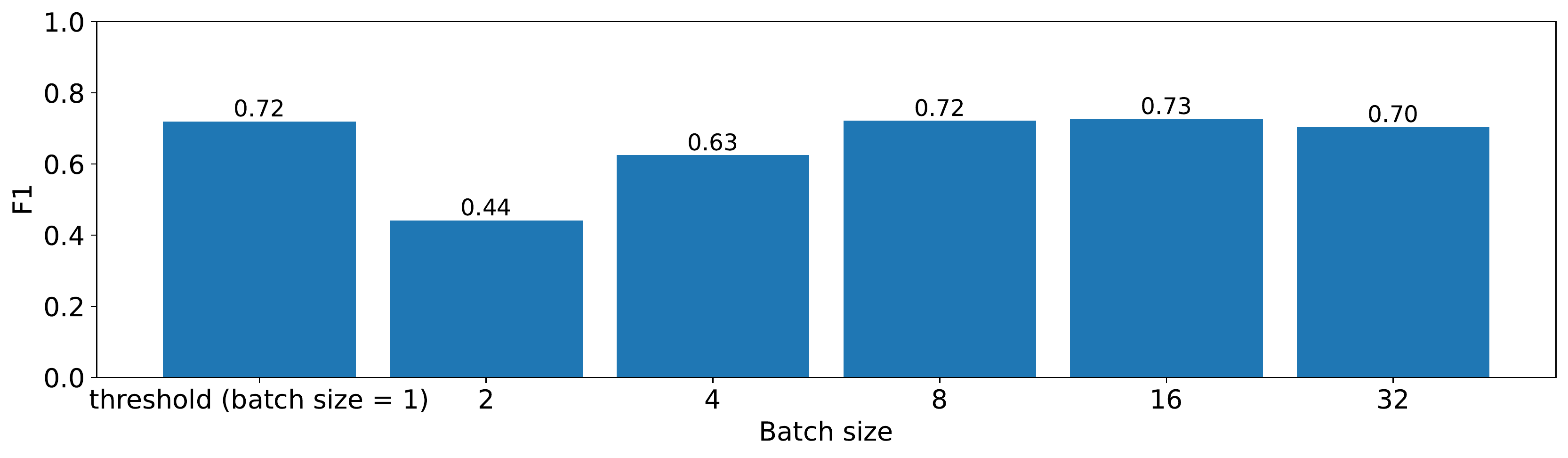}
\caption{Comparison of F1 scores using KS-test with different batch sizes. For batch size is 1, we use threshold on MSP with default value 0.9.}
\vspace{-1em}
\label{fig:f1_scores}
\end{figure}

\subsubsection{Chosen Detection Technique: Threshold on MSP}

After ruling out most of the techniques due to their lack of fit with on-device inference, we finalize a shortlist of two detection methods: the simple MSP threshold method that operates on one model-outputted logit vector at a time, and the KS-test statistical method, which operates on a batch of MSP scores. To evaluate these methods we measure their \emph{F1 score} to compare their effectiveness on an equal split of clean and drifted images (described in \S\ref{sec:detect-eval}) and a ResNet50 model (see \S\ref{sec:setup}) is used to generate the logit vectors. 
The F1 score is defined as: \begin{equation}\label{eq:F1}
F1 = 2 \cdot \frac{Precision \cdot Recall}{Precision + Recall} = 2 \cdot \frac{TP}{2TP + FP + FN}
\end{equation}
where a high score indicates that the detector has a good balance between precision and recall. In other words, the model is able to correctly identify true positives (the image is ``drifted'', \ie it belongs to a different distribution than the training set, and the detector thinks it is drifted) while minimizing false positives (the image is not drifted and the detector thinks it is drifted) and false negatives (the inverse). For the KS-test method, we test it with different batch sizes and assign the detection result (\ie a boolean value for drift or non-drift) \emph{on the whole batch}. For the threshold method, we set its default value to 0.9 (we evaluate this choice in \S\ref{sec:detect-eval}). 

Figure~\ref{fig:f1_scores} displays the results, 
showing that when the batch size is higher than 4, KS-test slightly outperforms the threshold method, but is worse when the batch size is lower.
However, batching results from device inference raises various tricky questions: should one batch inference results from a specific device over time? How long should we allow that time window to be? How do we batch results when we do not have sufficient samples from a single device to fit in one batch?

Since the threshold method performs very similarly to KS-test with large batches, and since using a batched detection method raises these thorny issues, we choose the threshold method with MSP as the default detection module in \name.

%% file: driftanalysis.tex
\subsection{Root Cause Drift Analysis} \label{sec:analysis}
As we will show later (\S\ref{sec:eval}), since the drift detection algorithm operates without any user input or labeled data, and it is simply a function of the model's uncertainty, the detection algorithm is somewhat noisy for each individual detection. Therefore, we need a more accurate mechanism at a system level to determine whether drift has actually occurred, and why. This subsection explores the design of this mechanism, which we call the \emph{root cause drift analysis}.

\begin{table}[t!]
\centering
\footnotesize
\begin{tabular}{|c|c|c|c|c|}
\hline
\textbf{Time} & \multicolumn{1}{|l|}{\textbf{Device ID}}  & \multicolumn{1}{|l|}{\textbf{Weather}} & \multicolumn{1}{|l|}{\textbf{Location}}  & \multicolumn{1}{|l|}{\textbf{Drift}} \\
\hline
06:02:01 & android\_42 & clear-day & Helsinki & False \\ 
\hline
06:02:23 & android\_21 & clear-day & New York & False \\ 
\hline
06:04:55 & android\_21 & clear-day & New York & True \\ 
\hline
08:03:32 & android\_21 & snow & New York & True \\ 
\hline
11:05:01 & android\_42 & snow & Helsinki & True \\ 
\hline
\end{tabular}
\caption{Example of drift log.}
\label{tab:log_example}
\vspace{-1em}
\end{table}

In \name, for each inference, the device sends to the cloud the \OOD detection results, as well as metadata about the device and its environment (\eg weather and geolocation, as depicted by the example in Table~\ref{tab:log_example}). Each entry is called the \emph{drift log entry}, and the entries are assembled in a single large database in the cloud, called the \emph{drift log}. Hence, the drift log is a live global view of \OOD across all devices. 

We walk through the algorithm using a toy example of two devices (one in New York, one in Helsinki) running an application for classifying animals in their natural settings. This application produces a drift log (Table~\ref{tab:log_example}). Each device generates a few drift entries, which contain metadata on the current location of the device and the weather it is operating in. 
This metadata can be either gleaned by the device itself (\eg its location) or generated by \name in the cloud via an external source (\eg for weather metadata, \name looks up the weather in a third-party weather API using the device's location).
In this example, the root cause of the drift is the snowy weather, which transforms the image sufficiently so that it diverges from the original training data -- imaged taken in clear days -- used to train the model. The example includes a false positive drift detection, \ie the third entry in the log, where the detection threshold algorithm marked the image as ``drifted'' (due to low confidence logit scores), even though the image was not. 
\vspace{-.5em}
\paragraph{Frequent itemset mining (FIM).}
To identify whether drifts exist and discover its root causes in the face of potential false positives and negatives, \name focuses only on clusters of drifts that are statistically significant. One possible approach is to directly apply an ML-based clustering algorithm on the entries flagged as drifts. 
We did not adopt this approach because clustering algorithms often require specifying how many clusters to divide the data into (\eg the $K$ parameter in K-means), and can be expensive, especially at scale~\cite{sculley2010web}. 

An alternative, classical data mining approach~\cite{agrawal1996fast,item-sets} is to explore whether sets of attributes in the drift log entries, \eg \{snow, New York\} in Table~\ref{tab:log_example}, frequently appear with the drift attribute. These attributes are termed in the data mining literature as \emph{frequent itemsets} and there exist many algorithms ~\cite{agrawal1996fast,914830,pei2000closet,borgelt2003efficient,han2000mining,borgelt2005implementation} for frequent itemsets mining (FIM).

\begin{table}[t!]
\centering
\footnotesize
\resizebox{\columnwidth}{!}{
\begin{tabular}{|c|c|c|c|c|c|c|c|}
\hline
\textbf{Rank} & \multicolumn{1}{|l|}{\textbf{Occurrence}}  & \multicolumn{1}{|l|}{\textbf{Support}} & \multicolumn{1}{|l|}{\textbf{Risk Ratio}}  & \multicolumn{1}{|l|}{\textbf{Confidence}}& \multicolumn{1}{|l|}{\textbf{Weather}} & \multicolumn{1}{|l|}{\textbf{Location}}& \multicolumn{1}{|l|}{\textbf{Location}}\\
\hline
0 & 0.4 & 0.67 & 3 & 1 & snow & - & -\\ 
\hline
1 & 0.2  & 0.3 & 2 & 1 & snow & - & android\_21\\
\hline
2 & 0.2  & 0.3 & 2 & 1 & snow & - & android\_42\\
\hline
3 & 0.2  & 0.3 & 2 & 1 & snow & New York & -\\
\hline
4 & 0.2  & 0.3 & 2 & 1 & snow & Helsinki & -\\
\hline
5 & 0.4  & 0.7 & 1.3 & 0.67 & - & - & android\_21\\
\hline
6 & 0.4  & 0.7 & 1.3 & 0.67 & - & New York & -\\
\hline
7 & 0.4  & 0.7 & 1.3 & 0.67 & - & New York & android\_21\\
\hline
8 & 0.2 & 0.33 & 0.75 & 0.5 &- & - & android\_21\\
\hline
9 & 0.2 & 0.33 & 0.75 & 0.5 & - & - & android\_42\\
\hline
10 & 0.2 & 0.33 & 0.75 & 0.5 &- & Helsinki & -\\
\hline
11 & 0.2 & 0.33 & 0.75 & 0.5 & clear-day & - & android\_21\\
\hline
12 & 0.2 & 0.33 & 0.75 & 0.5 & - & Helsinki & android\_42\\
\hline
13 & 0.2 & 0.33 & 0.75 & 0.5 & clear-day & New York & -\\
\hline
14 & 0.2 & 0.33 & 0.75 & 0.5 & - & Helsinki & -\\
\hline
15 & 0.2 & 0.33 & 0.33 & 0.33 & clear-day & - & -\\
\hline
\end{tabular}
}
\caption{Example of frequent itemset mining results.}
\vspace{-1em}
\label{tab:diff_result}
\end{table}

\begin{figure}[t!]
\centering
\includegraphics[width=\linewidth]{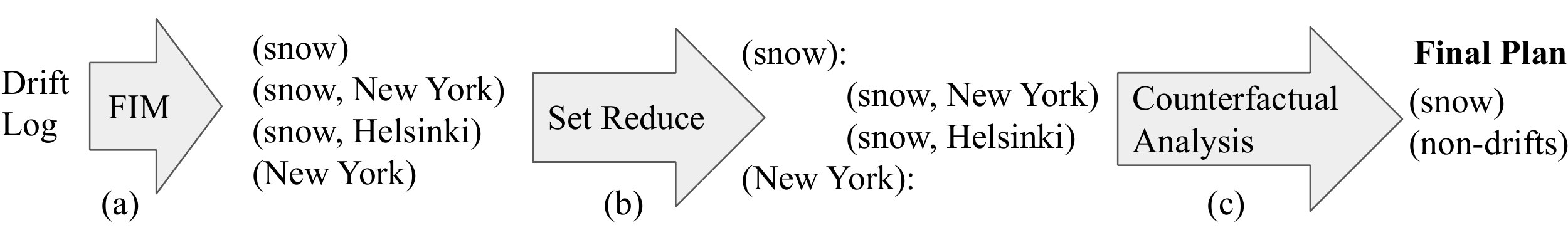}
\caption{Flow of root cause analysis.}
\vspace{-1em}
\label{fig:drift_analysis}
\end{figure}

In the first stage of root cause analysis, shown in Figure~\ref{fig:drift_analysis}(a), \name implements an FIM mechanism that employs the \emph{apriori algorithm}~\cite{apriori}. The apriori algorithm is a common method for FIM, which first identifies frequent individual attributes in the entry and then uses them to generate larger sets of attributes associated with the drift attribute. 
The algorithm calculates several metrics starting from sets containing each single attribute, and then the sets containing combinations of attributes which form subsets of the single ones, \eg \{snow, New York\} is a subset of \{snow\}. It filters the sets by checking if statistics from the metrics pass certain thresholds to consider a certain set as a cause of drift, since the drift detection is noisy. Finally, it ranks which sets of attributes are the leading causes of drift. The metrics and ranking for the example drift log in Table~\ref{tab:log_example} are shown in Table~\ref{tab:diff_result}. 

The \emph{occurrence} of a set of attributes measures how often it appears in the drift log. 
The \emph{support} of a set of attributes measures how often it is marked as ``drift'' as a proportion of all log entries that are marked as drifted. E.g., \{snow\} has a support of 0.67, because $\frac{2}{3}$ of the drift entries have the attribute ``snow''. 
The \emph{confidence} of an attribute set is how often it is marked as drift as a proportion of all its appearances in the table. 
Finally, the \emph{risk ratio} of a set measures how much likelier entries that are marked as drifted contain the set of attributes versus not containing it. E.g., for \{snow, Helsinki\}, the risk ratio is 2, because the probability of an entry being drift given it contains \{snow, Helsinki\} is 1 and the probability of an entry being drift given it does not contain \{snow, Helsinki\} is $\frac{1}{2}$. 

By default, \name uses the risk ratio to rank the results, because it measures the importance of a specific root cause.
Following prior work~\cite{diff}, \name sets the maximum number of attributes that can belong to a single root cause to 3, and uses values of 0.01, 0.01, 0.51 and 1.1 for minimum occurrence, support, confidence and risk ratio respectively, by default. 
\vspace{-.5em}
\paragraph{Practical limitations of FIM.}
While FIM is a useful technique to rank root causes and set some thresholds on which root causes are important, systems that rely on it typically assume that a human will then inspect the results and manually choose the appropriate root causes~\cite{diff}.
In our example in Table~\ref{tab:diff_result}, the top seven rows are all possible root causes, since they pass the four thresholds.
However, it would not be a good idea to simply adapt models to each one of these root causes, since they are overlapping and even contain subsets of each other (\eg \{snow, New York\} is a subset of \{snow\}). 
Therefore, \name needs a way to narrow down this set of possible root causes so it can apply model adaptations sparingly, covering as many relevant root causes as possible without requiring operator instruction.

\paragraph{Set reduction.}
There are two types of possibly redundant root causes that may have already been addressed in other root causes, and thus can be \emph{reduced}. First, a root cause which is a subset of other root causes that have broader coverage. For example, \{snow, New York\} is a subset of both \{snow\} and \{New York\}, so if one of these two is selected as root causes for model adaptations, there is no reason to select their intersection,\eg\{snow, New York\}. \name's set reduction algorithm merges the subset root cause into one of its super-sets that has higher ranking, \eg \{snow, New York\} is merged into \{snow\} instead of \{New York\}, because \{snow\} is ranked higher (rank 0). This stage is termed as \emph{set reduction} and is depicted in Figure~\ref{fig:drift_analysis}(b). 
\vspace{-.5em}
\paragraph{Counterfactual analysis.}
The second cause of redundancy is overlapping causes, when the entries with drift are already ``covered'' by some other root causes. 
In our example, $\frac{2}{3}$ entries from New York contain drift signals though $\frac{1}{3}$ are false positives. However, this subset of images is already addressed by a higher-ranked root cause, \eg \{snow\}. \name filters out this type of root cause by applying counterfactual analysis \cite{lewis1973counterfactuals}, which iteratively modifies the drift attribute from the log entries associated with a higher-ranked root cause to be ``false'', and tests whether the lower-ranked root cause is still statistically significant by checking the four metrics after these drift attributes have been modified. If it is still statistically significant, \name add this lower-ranked root cause into the final result. Note that the first top-ranked root cause is directly added to the final result. This iterative process is shown in Figure~\ref{fig:drift_analysis}(c) and it runs until all root causes from the set reduction output are exhausted. Eventually, \name groups images associated with each filtered root cause, \eg images that have ``snow'' in their meta-data. Note that \name also filters a set of images that are ``clean'' when they are not associated with previously discovered root causes. 
Algorithm~\ref{algo:drift_analysis} summarizes the root cause analysis mechanism.
\vspace{-.5em}
\paragraph{Limitations.}
The primary limitation of our root cause analysis is that the drift log attributes may not capture all possible root causes of drifts. For example, certain devices may have camera lenses from a particular manufacturer that are causing image corruption, but the drift log might not contain an attribute of the lens manufacturer for each device. In such cases, \name may group devices by their models, their location, or even by device ID, even though the ``real'' root cause is the lens. In such cases, \name should still automatically produce by-cause adaptations, which would capture these data drifts, but the ML operator team would have to manually investigate the issue in order to understand the real root cause. To combat this issue, in large-scale settings, the list of attributes should be exhaustive, to capture unforeseen causes of drift.


\begin{algorithm}[t]
\footnotesize
	\caption{Root cause analysis algorithms} 
	$FIM():$\\
	{\bf Output}: Extracts frequent sets of attributes associated with drift.\\\\
 $Set\_Reduction(FIM\_table):$\\
    {\bf Input}: Sorted list of attribute sets.\\
    {\bf Output}: Mapping between the most coarse-grained attribute sets and their subsets. Ties between coarse-grained sets are broken by ranking.\\\\
$Passes\_Drift\_Threshold(items):$
\\
    {\bf Input}: Attribute set potentially causing drift.\\
    {\bf Output}: True if the set passes thresholds (occurrence, support, confidence, and risk). False otherwise.\\

\textbf{Drift Analysis:}
\begin{algorithmic}[1]

\State $FIM\_list \gets FIM()$

\State $coarse\_associations~\gets~Set\_Reduction(FIM\_list)$

\State $root\_causes \gets [\ ]$
\While{$coarse\_associations$}
    \State $curr\_coarse\_cause \gets coarse\_associations.pop()$
    \If{$Passes\_Drift\_Threshold(curr\_coarse\_cause)$} 
        \State $root\_causes.append(curr\_coarse\_cause.key)$
        \State $Mark\_No\_Drift(curr\_coarse\_cause.key)$
    \Else
        \ForEach{$subset$ in $curr\_coarse\_cause.value$}
            \If{$Passes\_Drift\_Threshold(subset)$}
                \State $root\_causes.append(subset)$
            \EndIf
        \EndFor    
        
    \EndIf 
    
\EndWhile

\State \textbf{return} $root\_causes$
\end{algorithmic} 
\label{algo:drift_analysis} 
\end{algorithm}
\vspace{-1em}

%% file: adaptation.tex
\subsection{By-Cause Adaptation} \label{sec:adaptation}
Most model adaptation techniques~\cite{
li2017learning,rebuffi2017icarl,shmelkov2017incremental,yin2018feature,maltoni2019continuous} and end-to-end ML systems that incorporate adaptation (\eg SageMaker~\cite{sagemaker}, TFX~\cite{TFX}) require labeled data, which are unsuitable for our design. 

We focus on two main self-supervised methods for adapting models to \OOD, motivate the idea of adapting a model to a \emph{specific cause of drift}, and introduce how Nazar chooses different adapted model versions for inference. 

\paragraph{Self-supervised adaptation.}
\name's model adaptation builds on prior work on self-supervised adaptation. The intuition is to select a self-supervised objective (\eg minimize the entropy of the model output) that captures some invariant of the data itself without requiring any labels. Upon data drift, this objective is no longer optimal, therefore we can optimize the model weights to minimize this objective again, effectively adapting the model to the drift.

We consider two methods: 
TENT~\cite{DBLP:journals/corr/abs-2006-10726} and MEMO~\cite{DBLP:journals/corr/abs-2110-09506}. 
TENT adapts any pre-trained probabilistic model by minimizing the entropy on the logits output by the model, on a batch of images. 
MEMO adapts on marginal entropy, which is the entropy of the averaged output logits, over different augmented copies of a single image, \eg by rotating and posterizing the image. The idea is that the model should be (a) invariant across augmented versions of the image, and (b) confident in its predictions even for heavily-augmented versions of the image. 

We make three modifications to these methods to make them practical for the setting of large ML deployments on mobile devices that \name targets. First, both methods tightly couple adaptation and inference: they adapt the models based on the current inference inputs, and then immediately run inference on the inputs. However, as discussed in \S\ref{sec:background}, many mobile devices are too wimpy for on-device adaption, so we decouple adaption and inference in \name. Second, MEMO adapts a whole model including all its layers, significantly increasing network consumption and model storage costs because each adaptation leads to a whole new version of the model weights. We thus modify MEMO to adapt only the \emph{batch normalization (BN) layer}~\cite{BatchNorm}, similar to TENT. As an example, in ResNet50 the BN layer is $217\times$ smaller than the full model (0.4MB vs. 92MB). Our experiments show that adapting only the BN layer in MEMO has the same accuracy boost as adapting all layers. Third, we modify MEMO to adapt based on a small batch of inputs instead of on every input because the latter would be impractical in our setting, incurring too frequent adaptations even for one-off or falsely-detected drifts.





\paragraph{Benefit of adapting by cause over adapting blindly.}
Prior adaption methods unrealistically assume a single cause behind the \OOD and adapt on all inputs, but in practice \OOD can originate from different sources: for example, a traveler may bring her device from location to location over a short period of time, experiencing very different input distributions each corresponding to a different cause (location). We show in our experiments below that adapting by cause improves model performance significantly over adapting blindly on all inputs. Moreover, to illustrate the need for accurate root cause analysis, we show that a model adapted to one cause of \OOD has poor performance when applied to data that is drifted due to a different cause or non-drifted (\ie clean) data.


\begin{table}[t!]
\centering
\footnotesize
\begin{tabular}{|l|r|}
\hline
Methods & \multicolumn{1}{|l|}{Average Accuracy (\%)} \\
\hline
No-adapt & 38.7 \\
\hline 
By-cause (TENT)& 61.5 \\
\hline
By-cause (MEMO) & 42.3\\
\hline
Adapt-all (TENT) & 42.4\\
\hline
Adapt-all (MEMO) & 30.3\\
\hline
\end{tabular}
\caption{TENT and MEMO when adapting them by-cause, and using a single model to adapt to all sources.}
\vspace{-1em}
\label{tab:adaptation-comparison}
\end{table}
We adapt a pre-trained ResNet50 on a dataset that contains 16 different types of data drifts, such as weather-based corruptions and image blur (\S\ref{sec:eval} describes the evaluation setup), and a portion of clean data. We run TENT and MEMO in two settings: (a) adapt and test a model for each cause of drift and clean data (17 total models), termed \emph{by-cause}, and (b) adapt one model on a mix of all 16 sources of drifts and clean data, and test them on the same test set as in (a), termed \emph{adapt-all}. We also evaluate non-adapted ResNet50 on the same test set. 

The average test accuracy is shown in Table~\ref{tab:adaptation-comparison}. Adapting on data by-cause using both methods improves accuracy over the original model, by 22.8\% for TENT and 3.6\% for MEMO; whereas adapting on all the data degrades accuracy with MEMO and leads to small gains with TENT. The reason is that adapting models to multiple divergent distributions without supervision can cause models to underfit and leads to low performance. To further demonstrate this, we run an experiment where we adapt a model by-cause (foggy weather), and test it on images with other sources of drift. On average, the model obtains an average accuracy of only 16.4\% when run on images with other sources of drift, and an accuracy of 66.7\% on its own test set. Its accuracy on clean data is also very low: 26.8\%, while the model adapted on clean data has an accuracy of 74.6\% on the same clean test set.

Therefore, we conclude that \name has to adapt its models sparingly and apply them carefully; it should apply adapted models only when there is a true prevalent source of data drift that affects some portion of user devices, and ideally run the resulting adapted model on data that experiences the same source of drift.
These observations motivate \name to adapt different models by distinct root cause, so that the distribution of inference data will be as close as possible to the one of the training data. Ideally, each by-cause adapted model should run inference on data that belongs to the same distribution it was adapted on, while a continuously adapted ``clean'' model should be run on clean data. Since no ground truth exists, \name has to rely on its root cause analysis to determine which devices are experiencing \OOD and from which source.
In addition, this motivates us to use TENT as the default method to adapt the models in \name, since TENT largely outperforms MEMO in the both strategies in our experiment, as shown in Table~\ref{tab:adaptation-comparison}. However, ML operators are free to use any self-supervised adaptation method with \name.






\paragraph{Consolidating model versions.}
Adapting by-cause generates multiple model versions. We cannot allow the number of models to grow unbounded, especially on resource-constrained devices. \name constrains the number of models that can be stored on each user device, by periodically consolidating them. To consolidate different versions, \name keeps a global view of the model pool, in a least recently updated (LRU) list, which contains the sets of models to be deployed. The oldest models make room for the newer ones. Orthogonally to the LRU algorithm, if a new model version belongs to the exact same set of attributes as an existing one, the older one is evicted without having to evict the tail of the LRU. In addition, if an incoming model version has a root cause that is a superset of an older model version, the older version gets evicted (similar to the set reduction algorithm in \S\ref{sec:analysis}). 

\paragraph{Picking which model to use for inference.}
During inference on the device, \name will use the most-recently updated model that has the highest number of matching attributes for inference, \eg \{rain, New York\} has more attributes matching than \{rain\} if the input image is associated with \{rain, New York\}. It uses the risk ratio ranking of a root cause, described in \S\ref{sec:analysis}, to break ties between model versions. If no matching model is found, it uses its ``clean'' model for inference. Note that model selection for inference is run on the device without any involvement from the cloud. 


%% file: implementation.tex
\section{Implementation} 

We now describe our prototype implementation of \name on Amazon AWS. To support a large number of devices and models, our implementation uses highly-scalable AWS services. Such services exist on all cloud providers, and our implementation could be easily ported to other clouds.

\paragraph{Drift log.}
We run the drift log on Amazon Aurora~\cite{aurora}, a relational database that supports over 120,000 writes per second and can be scaled to 128 tebibytes.
The drift log is stored as a table, where each row contains the metadata and whether the inference was detected as a \OOD. 

\paragraph{Root cause analysis and adaptations.}
The root cause analysis is run periodically as an Amazon Lambda function, which is triggered either automatically based on a configurable time window or manually by the ML operator. 
The FIM algorithm is run as a set of SQL queries on the drift log. 
The initial phase of the algorithm requires extracting the attributes frequently associated with drift. 
This can be implemented using a simple SQL \textsc{Count} aggregation, with appropriate conditions.
The function then calculates the FIM metrics, which are used to filter and rank the root causes. 
The function then runs the set reduction and counterfactual analysis.
Finally, each narrowed-down root cause is sent to a GPU-equipped instance, which conducts the adaptation by-cause.

%% file: evaluation.tex
\section{Evaluation}
\label{sec:eval}
This section answers the following questions:
\begin{denseitemize}
\item[{\bf Q1:}] How well does \name's detection algorithm detect different types of \OOD? (\S\ref{sec:detect-eval})
\item[{\bf Q2:}] How effective is \name's root cause analysis in identifying the root cause of drift? (\S\ref{sec:analysis-eval})
\item[{\bf Q3:}] How does by-cause adaptation perform compared to prior approaches? (\S\ref{sec:adapt-eval})
\item[{\bf Q4:}] Does the detector evolve with model adaptation? (\S\ref{sec:interaction-eval})
\item[{\bf Q5:}] What is the accuracy gain in end-to-end workloads? (\S\ref{sec:end-to-end})
\item[{\bf Q6:}] How quickly does \name adapt to \OOD and how well does the root cause analysis scale? (\S\ref{sec:implementation-eval})
\end{denseitemize}

We first describe our datasets (\S\ref{sec:datasets}) and our experimental setup (\S\ref{sec:setup}), and then answer the evaluation questions.

\subsection{Datasets}
\label{sec:datasets}
\begin{figure}[t!]
  \centering
      \begin{subfigure}[b]{\columnwidth}
    \includegraphics[width=0.24\linewidth]{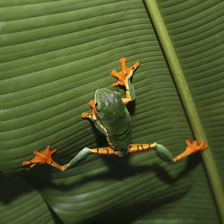}
    \includegraphics[width=0.24\linewidth]{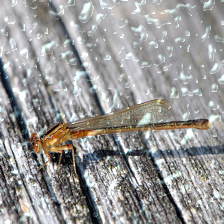}
    \includegraphics[width=0.24\linewidth]{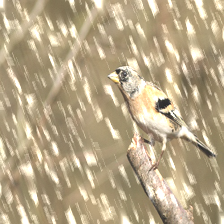}
    \includegraphics[width=0.24\linewidth]{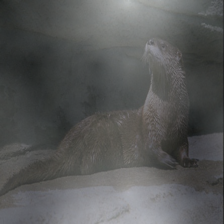}
    \subcaption{Wildlife animal example.}
    \end{subfigure}
  \medskip
    \begin{subfigure}[b]{\columnwidth}
    \includegraphics[width=0.24\linewidth]{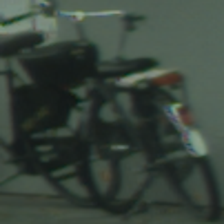}
    \includegraphics[width=0.24\linewidth]{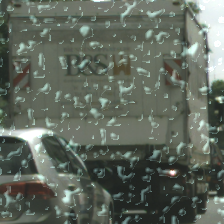}
    \includegraphics[width=0.24\linewidth]{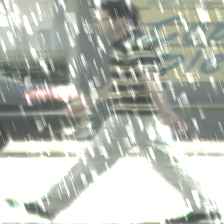}
    \includegraphics[width=0.24\linewidth]{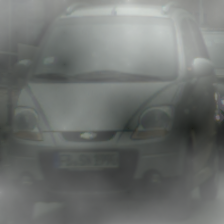}
 \subcaption{Cityscapes example.}
 \vspace{-1em}
  \end{subfigure}
  \caption{Examples of clean image and images with rain, snow and fog (corruption severity of 3) from the two datasets.}
  \vspace{-1em}
  \label{fig:weather_comparison}
\end{figure}
\input{datasets}
\begin{figure*}[t!]
\small
\centering
\begin{subfigure}[t]{0.33\textwidth}
\includegraphics[width=\textwidth]{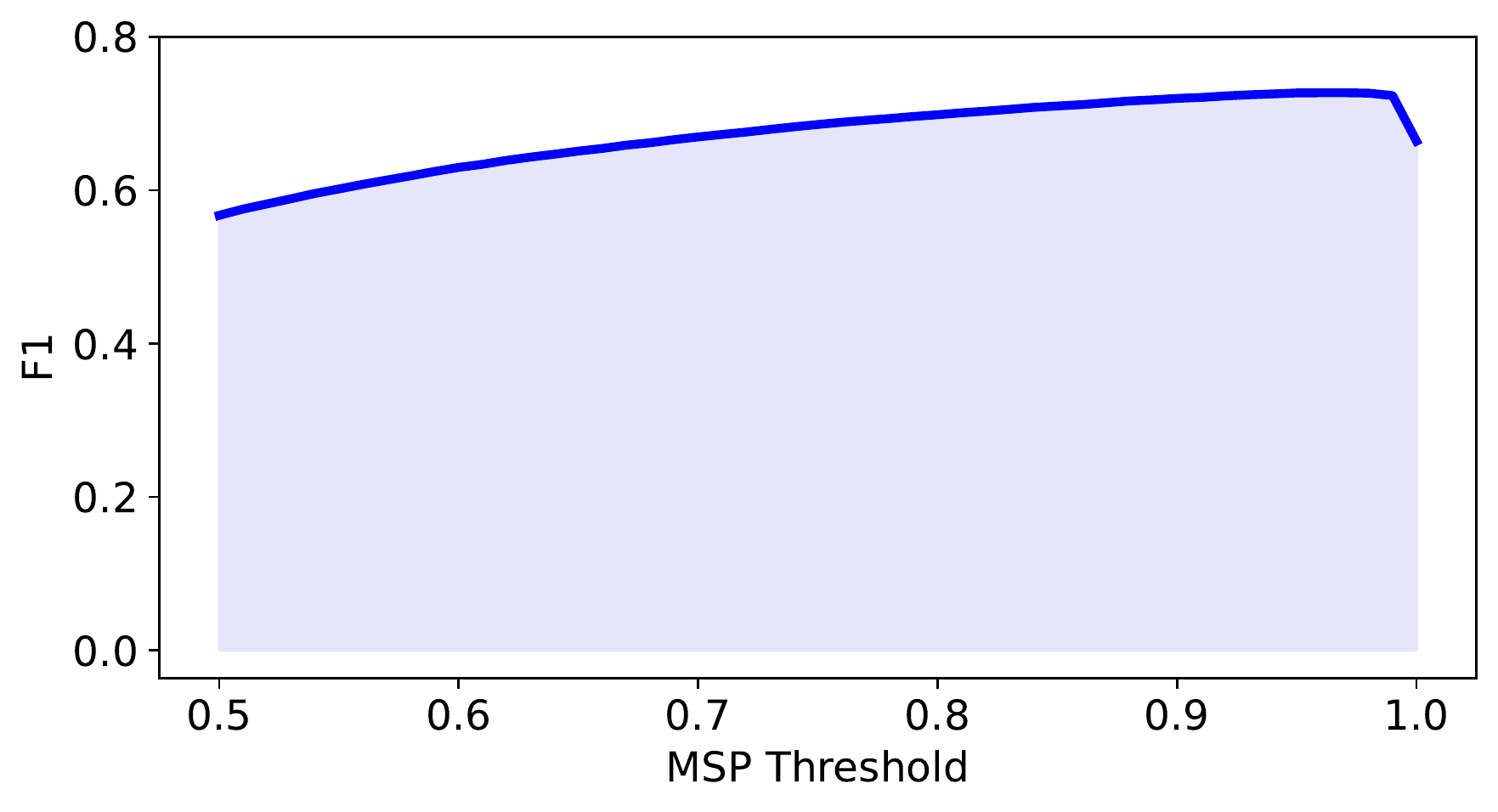}
\caption{F1 score by MSP threshold.}
\label{fig:detection_thresholding}
\end{subfigure}~
\begin{subfigure}[t]{0.34\textwidth}
\includegraphics[width=\textwidth]{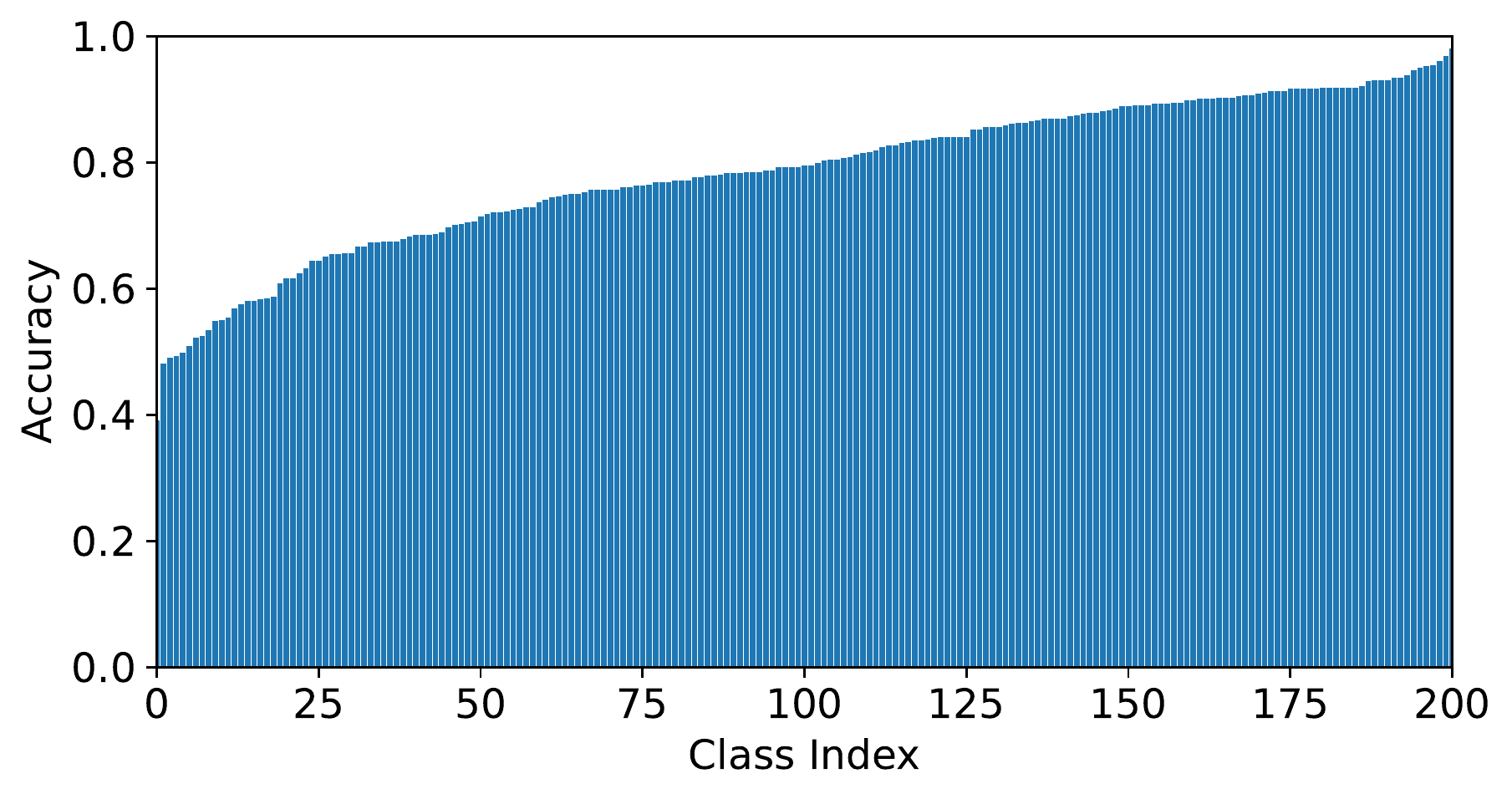}
\caption{Accuracy across classes.}
\label{fig:sorted_201cls}
\end{subfigure}~
\begin{subfigure}[t]{0.32\textwidth}
\includegraphics[width=\textwidth]{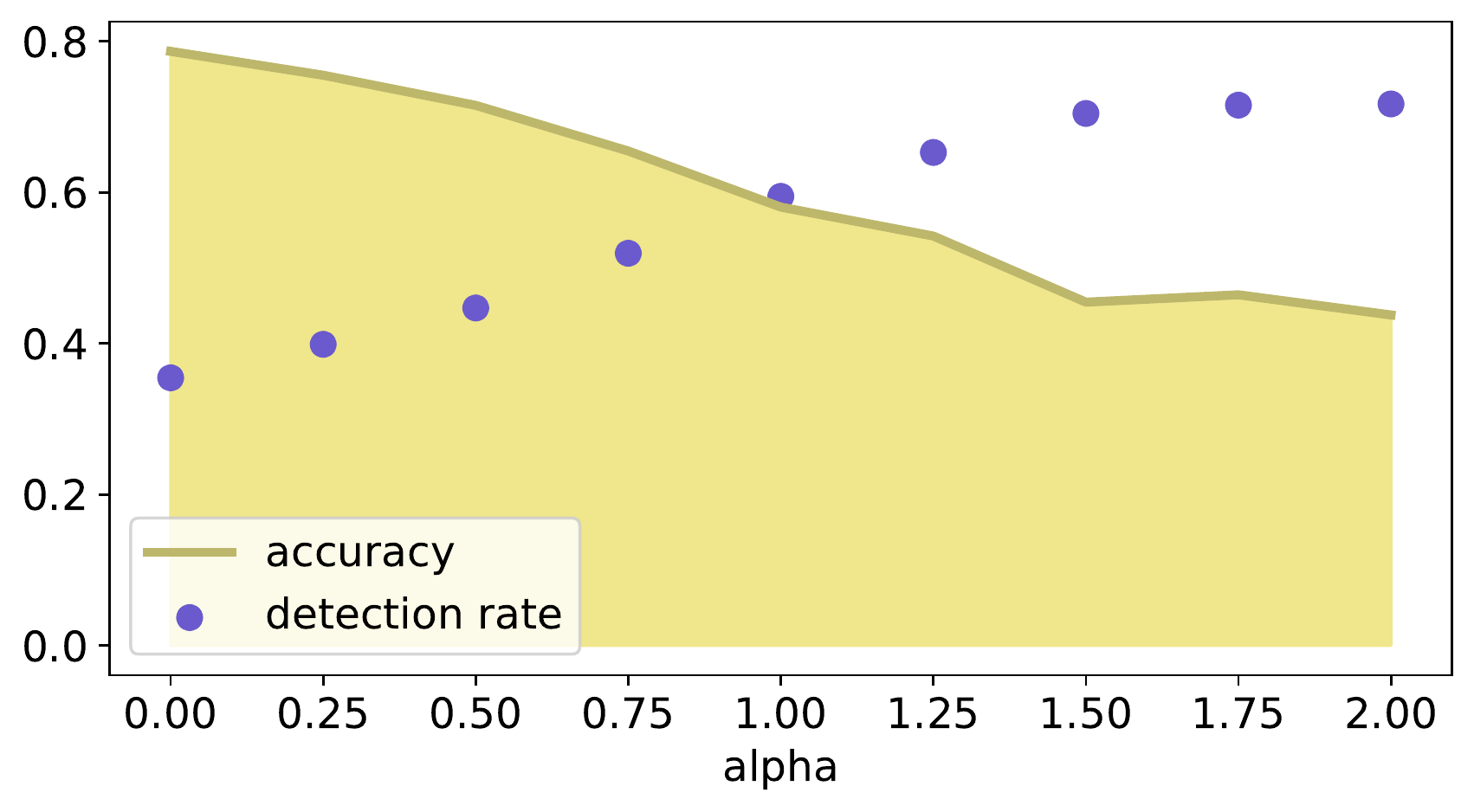}
\caption{Rate of detection and accuracy vs skew.}
\label{fig:detecting_skewness}
\end{subfigure}
\caption{\ref{fig:detection_thresholding} shows F1 scores of MSP detector are fairly stable on different threshold values. \ref{fig:sorted_201cls} shows average accuracy on different animal classes is highly variable. \ref{fig:detecting_skewness} shows the accuracy decreases and detection rate increases when skew is high.}
\label{fig:detection}
\end{figure*}

\begin{table*}[t!]
\centering
\small
\resizebox{\linewidth}{!}{
\begin{tabular}{|l|r|r|r|r|r|r|r|r|r|}
\hline
Analysis Methods / Ground Truth Drifts & \multicolumn{1}{|l|}{None} & \multicolumn{1}{|l|}{Rain} & \multicolumn{1}{|l|}{Snow} & \multicolumn{1}{|l|}{Fog} & \multicolumn{1}{|l|}{Fog \& Snow}& \multicolumn{1}{|l|}{Fog \& Rain}& \multicolumn{1}{|l|}{Snow \& Rain}&
\multicolumn{1}{|l|}{Snow \& Rain \& Fog}\\
\hline
FIM  &1 & 0.919 & 0.773 & 1 & 0.891 & 0.921 & 0.926 & 0.917 \\
\hline 
FIM with Set Reduction &1 & 0.919 & 0.773 & 1 & 0.891 & 0.921 & 0.934 & 0.919 \\
\hline 
FIM with Set Reduction and CF Analysis &1&  1&  0.874&  1& 1&  1&  1&  1  \\
\hline
\end{tabular}}
\caption{Evaluation of root cause analysis algorithms with Fowlkes-Mallows Score (1 is optimal).}
\vspace{-1em}
\label{tab:FMS}
\end{table*}

\subsection{Evaluation Setup}
\label{sec:setup}

\paragraph{AWS setup.} Our AWS implementation uses the following configurations. The drift log is run on Amazon Aurora~\cite{aurora} with a 5.7.mysql\_aurora.2.11.1 engine on a db.r6g.2xlarge instance~\cite{awsrds}. The drift analysis is run as an AWS Lambda~\cite{awslambda} serverless function with 256MB of memory. The adaptation is run on a p3.2xlarge EC2 instance, which is equipped with an NVIDIA Tesla V100 GPU~\cite{awsec2p3}. 

\paragraph{Models.}
We use three image classification models in all experiments, ResNet18, ResNet34 and ResNet50, which are commonly-used model architectures for mobile devices~\cite{torchmobile}. Each model is trained from scratch until convergence. 
The accuracy of each model on clean (non-drifted) validation dataset with cityscapes is 83.6\%, 83.9\% and 83.7\%, respectively, while the accuracy of the base model on the animal dataset is 72.1\%, 75.4\% and 76.1\%. 

\paragraph{Data drifts.}
Prior work on drifts apply 16 types of different data drift on ImageNet~\cite{ImageNet-C}. In microbenchmarks in \S\ref{sec:detect-eval}, \S\ref{sec:adapt-eval} and \S\ref{sec:interaction-eval} we show how \name handles all 16 types of drifts. Our end-to-end evaluation (\S\ref{sec:end-to-end}) contains only the three weather-related drifts (snow, rain, fog), as the distribution of weather is dynamic and driven by real historical weather records. 29\% and 36\% of days in the cityscape and animal datasets, respectively, experience weather-related drifts. All 16 drifts use a severity level of 3 (out of 5) by default.

\paragraph{Baselines.}
We compare \name with two other settings: adapt-all, which is the baseline approach (\eg employed by Ekya, TENT, MEMO), where one model is continuously adapted on all input during each adaptation window, and no-adapt, which is a pre-trained model that is never adapted. Note that besides adapt-all, we also conducted experiments on TENT to adapt on only the images whose drift attribute is ``True'' but this method always yields worse performance than adapt-all and thus we do not present its results here.

\subsection{Detection (Q1)}
\label{sec:detect-eval}
To show how well the MSP threshold detects drifts, we measure the F1 scores calculated with different MSP thresholds on the logit outputs from Resnet50. 
We evenly applied 16 types of drifts on half of the streaming images in the animal dataset, as the positive set, and leave another half non-drifted, as the negative set.  We show the result in Figure~\ref{fig:detection_thresholding}, where the F1 score of MSP detector steadily increases until it reaches 0.73 and then decreases. 
The detector's sensitivity to threshold changes is small around a threshold of 0.9, which we adopt as a default value.

Further illustrating this point, in Figure~\ref{fig:detecting_skewness}, we vary the class skew, \ie we vary the distribution we sample from different classes, where under high skews we are much likelier to sample specific animal classes than others. The result shows that the detector rate significantly increases and the total accuracy degrades under high class skew.

These results indicate that the MSP threshold is a good indicator of different sources of drifts.

\subsection{Root Cause Analysis (Q2)}
\label{sec:analysis-eval}
We evaluate \name's root cause analysis using the Fowlkes-Mallows Score (FMS)~\cite{FMI}, which measures the similarity between two sets of clustering results. In our case, the two sets are ground truth root causes of drift, and ones that are identified by our root cause analysis.
The FMS score is:
\begin{equation}
    FMS = \sqrt{\frac{TP}{TP+FP}\cdot\frac{TP}{TP+FN}}
\end{equation}
\label{FMS}
where TP (true positive) is the number of pairs of data points that are clustered together in both sets, FP (false positive) is the number of pairs that are clustered together in the second set but not in the first, and FN (false negative) is the inverse of FP. The score ranges from 0 to 1, with higher values indicating more similarity between the clusters. We use 8 drift scenarios, depicted in Table~\ref{tab:FMS}, whose ground truth root causes of drifts are different combinations of three weather corruptions on the animal dataset for 14 days starting on January 1st, 2020, on a ResNet50. As an example, if the ground truth root causes are ``fog'' and ``snow'' then we apply only these two types of drifts on the dates that have foggy and snowy weather for each location but ignore ``rain'' drift on the rainy days. 

The combination of FIM, set reduction, and counterfactual analysis always yields the highest score under each scenario. In fact, it obtained the optimal FMS score under all drift scenarios except ``snow''.

\subsection{Adaptation (Q3)}
\label{sec:adapt-eval}
To evaluate whether by-cause adaptation performs better than the naive approach, \ie adapting on all coming images, 
%
we evenly split the images in the animal dataset in 17 partitions, 16 of which are transformed with different drifts, and one is left as clean, and run them on ResNet50.
To isolate the performance of the adaptation mechanism, we assume perfect knowledge of the underlying root cause of the \OOD, and that we adapt only to images with that root cause (\ie we assume \name's detector and root cause analysis are perfect). 
We treat the clean data as its own data source, and run an adaptation model for it as well.



\begin{figure}[t!]
\centering
\includegraphics[width=\columnwidth]{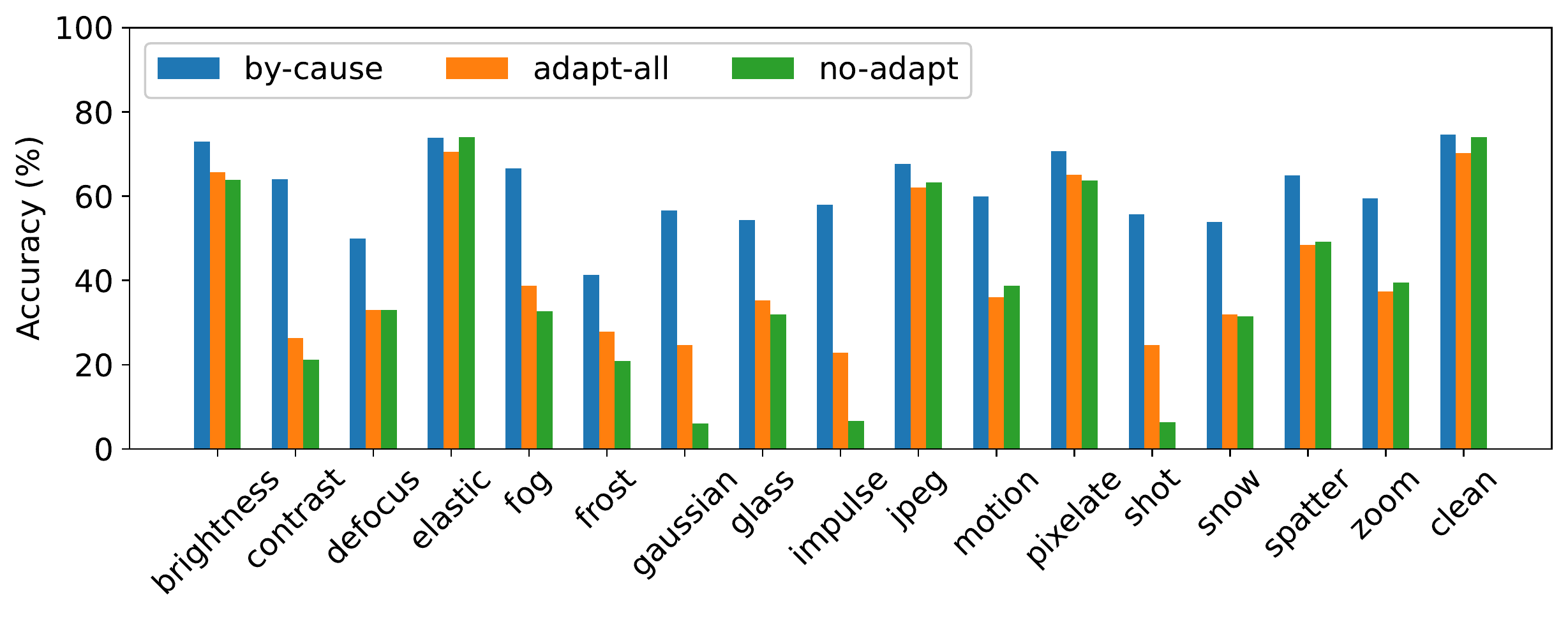}
\caption{Experiments of by-cause, adapt-all, and no-adapt on 16 different drifted data and clean data} 
\label{fig:adaptation-experiment}
\vspace{-1em}
\end{figure}

Figure~\ref{fig:adaptation-experiment} breaks down the average accuracy by drift type.
By-cause adaptation significantly and consistently outperforms the adapt-all. 
In addition, it provides a significant improvement compared to the baseline non-adapted model, while adapt-all sometimes degrades the accuracy of the non-adapted model.
The total average accuracy increase of by-cause adaptation in this experiment is very significant: it achieves a 61.5\% accuracy, compared to only 42.4\% for adapt-all and 38.7\% for the non-adapted model.

We conclude that by-cause adaptation shows promise in overcoming the limitations of traditional supervised adaptation approaches, paving the way for more scalable and efficient adaptation in real-world applications.

\begin{figure}[t!]
\centering
\includegraphics[width=\columnwidth]{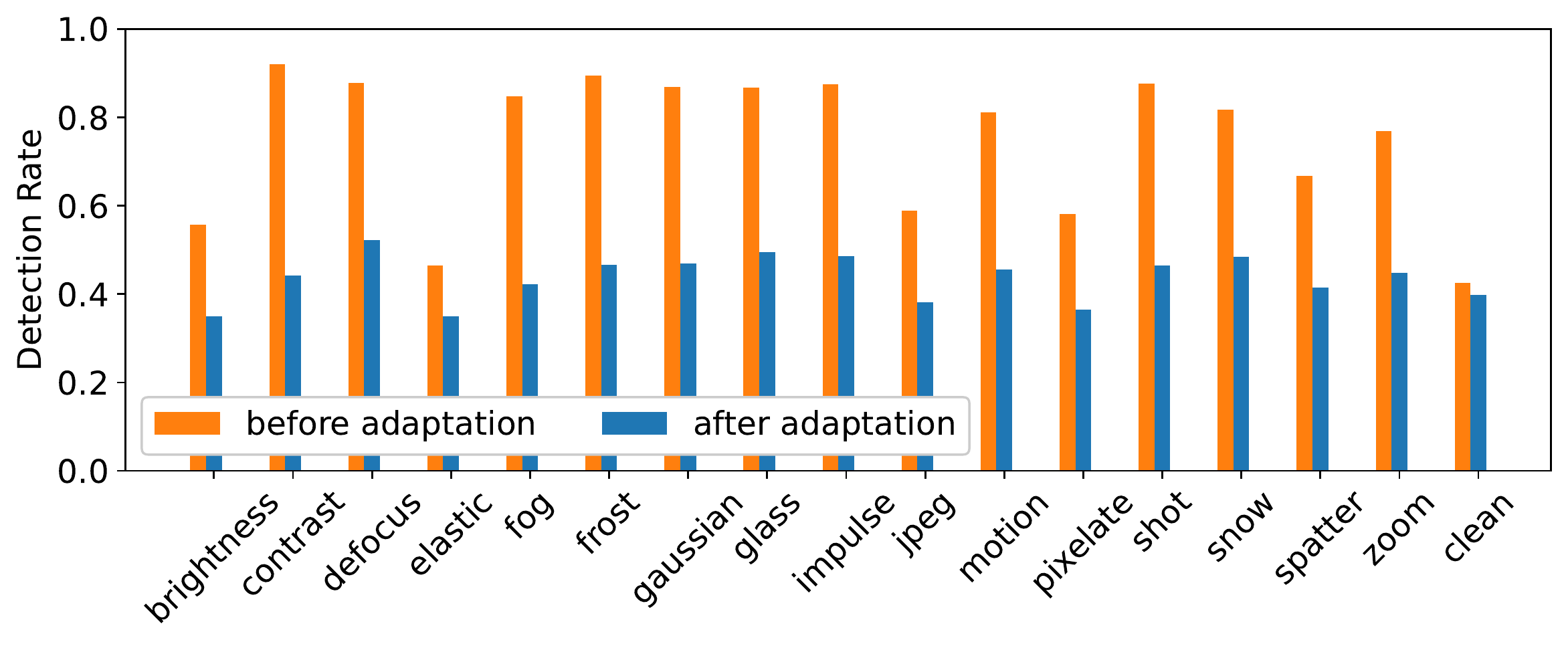}
\caption{Detection rate before and after adaptation.}
\vspace{-1em}
\label{fig:detection-rate-before-after}
\end{figure}

\begin{figure*}[ht!]
\captionsetup{font=footnotesize}
\centering
\begin{subfigure}[t]{0.244\textwidth}
\includegraphics[width=\textwidth]{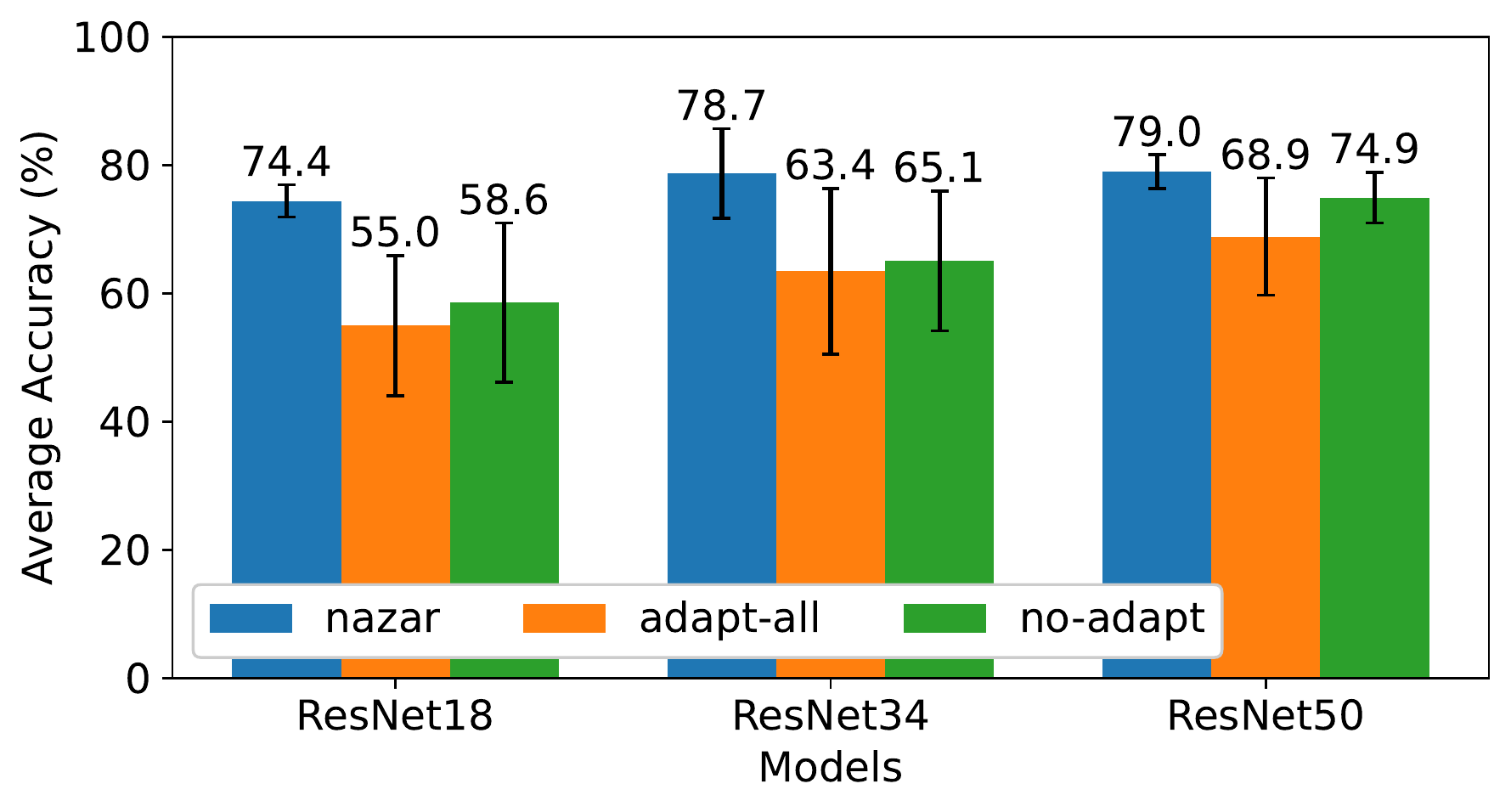}\
\subcaption{Average accuracy.}
\label{fig:cityscapes_8_avg}
\end{subfigure}~
\begin{subfigure}[t]{0.244\textwidth}
\includegraphics[width=\textwidth]{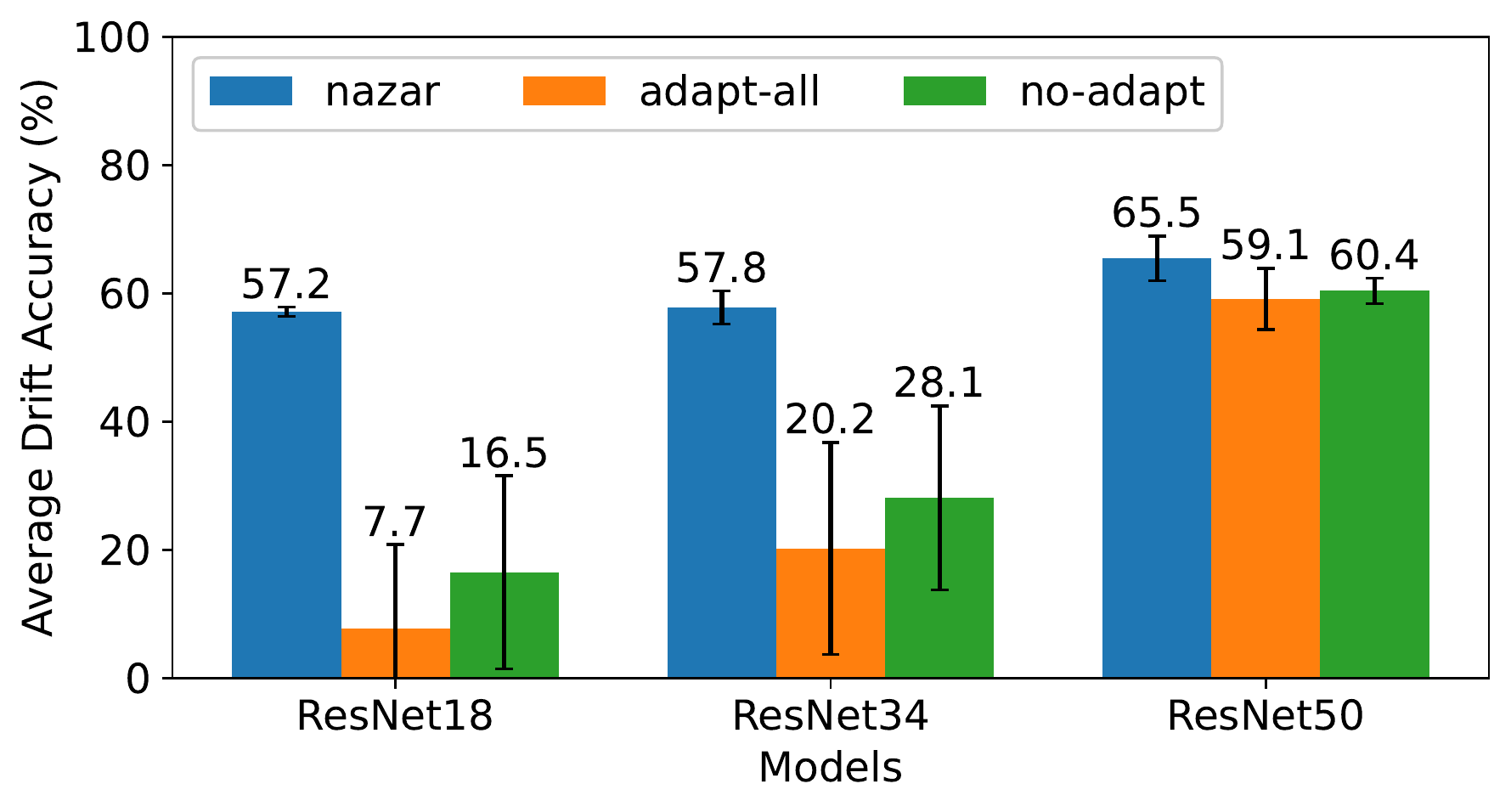}
\subcaption{Average accuracy of drifted data.}
\label{fig:cityscapes_8_corr}
\end{subfigure}~
\begin{subfigure}[t]{0.244\textwidth}
\includegraphics[width=\textwidth]{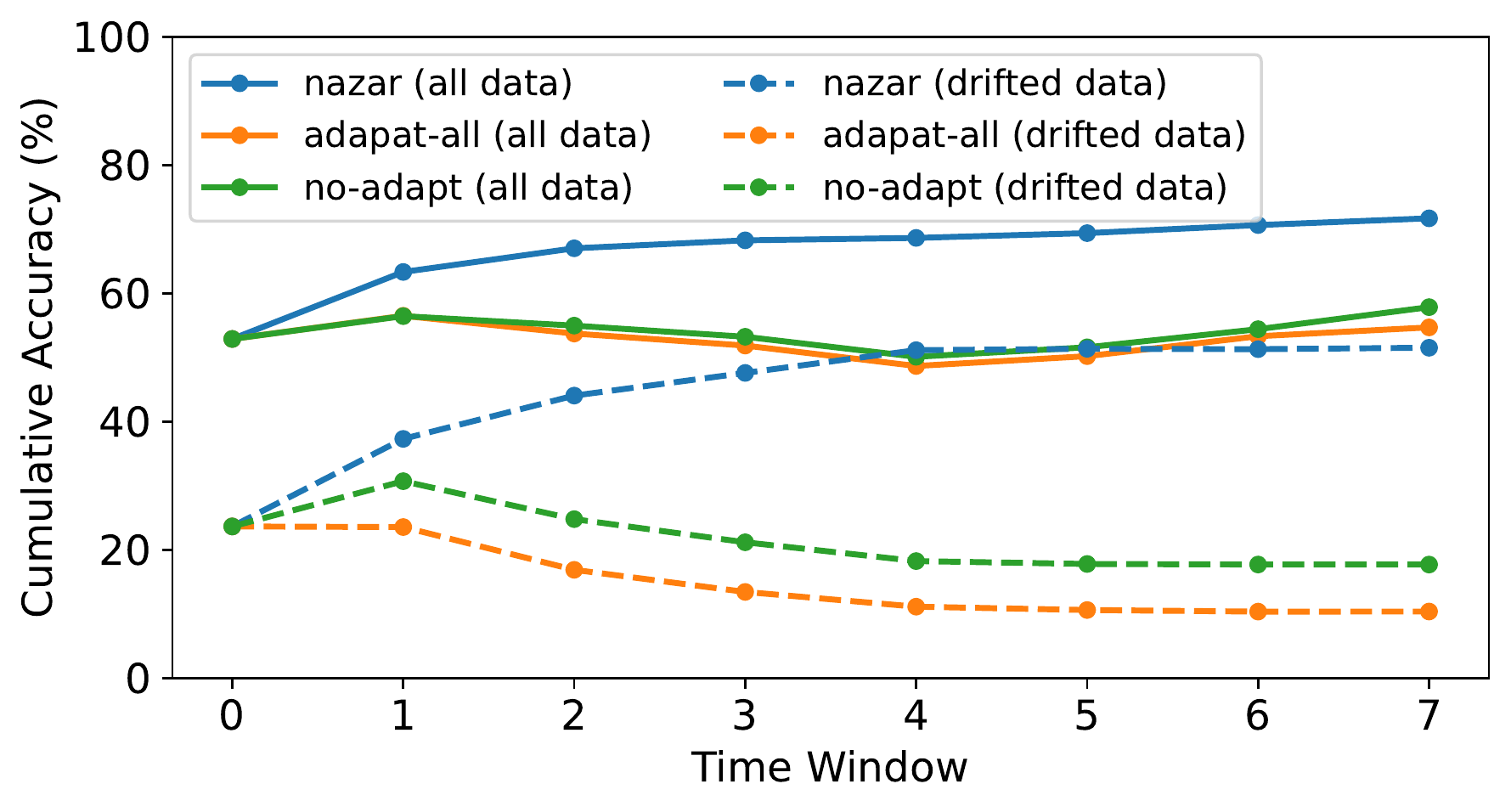}
\subcaption{ResNet18 accuracy over time.}
\label{fig:cityscapes_8_trace}
\end{subfigure}
\begin{subfigure}[t]{0.243\textwidth}
  \includegraphics[width=\textwidth]{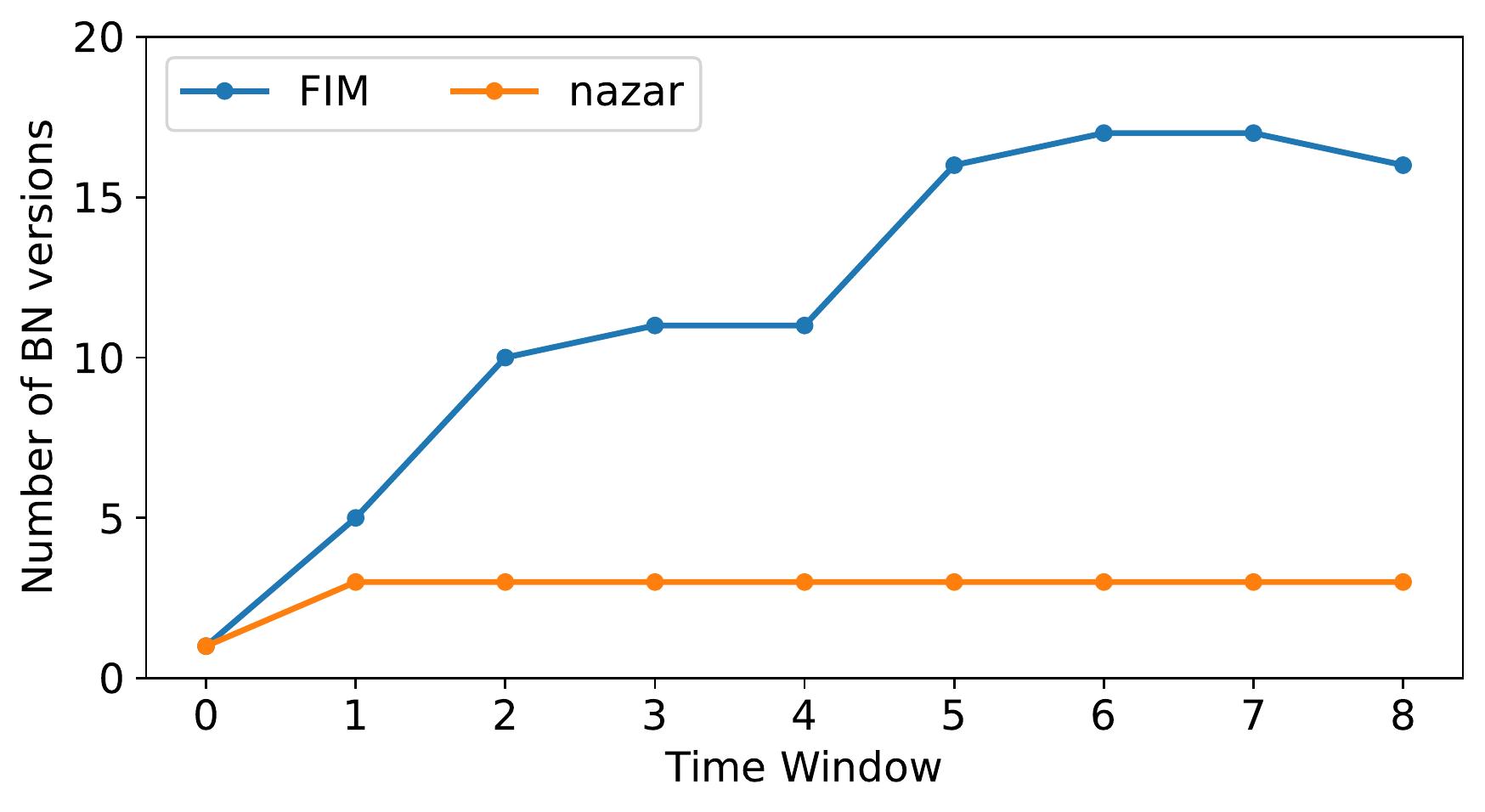}
  \subcaption{Number of model versions over time.}
  \label{fig:bn_verions}
\end{subfigure}
\caption{Results of experiments on cityscapes dataset. \ref{fig:cityscapes_8_avg} and \ref{fig:cityscapes_8_corr} show the average accuracy on all data and drifted data under three strategies. \ref{fig:cityscapes_8_trace} shows the trace of accumulated accuracy per time window. \ref{fig:bn_verions} shows the number of model versions on user devices when set reduction and counterfactual analysis are removed from \name.}
\vspace{-1em}
\end{figure*}
\label{fig:end-2-end_cityscapes_8}

\begin{figure*}[ht!]
\captionsetup{font=footnotesize}
  \centering
  \begin{subfigure}[t]{0.246\textwidth}
  \includegraphics[width=\textwidth]{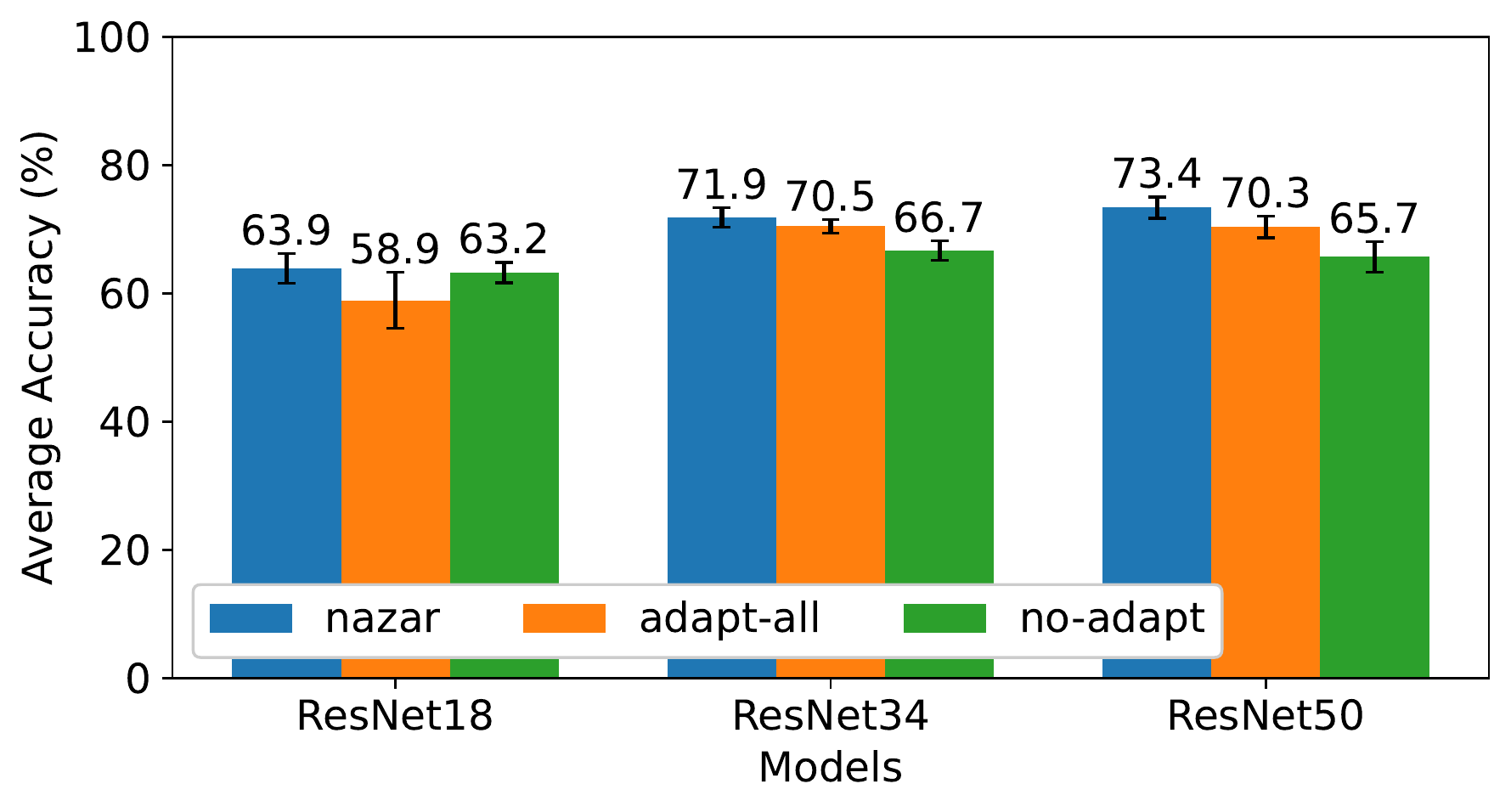}
  \subcaption{Average accuracy (S = 3).}
  \label{fig:avg_acc_s3}
  \end{subfigure}
  \begin{subfigure}[t]{0.246\textwidth}
  \includegraphics[width=\textwidth]{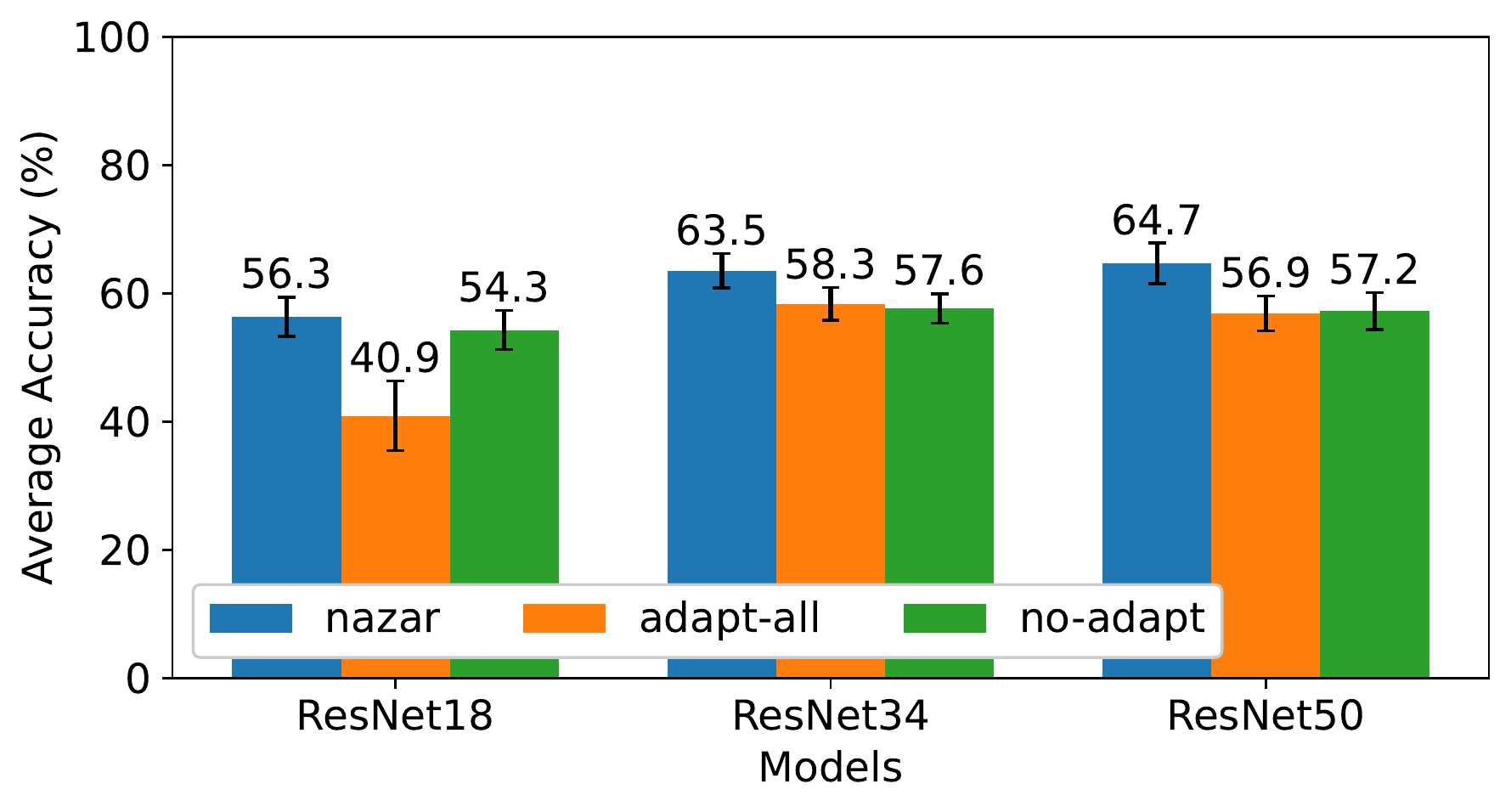}
  \subcaption{Average accuracy (S = 5).}
  \label{fig:avg_acc_s5}
  \end{subfigure}
  \begin{subfigure}[t]{0.246\textwidth}
  \includegraphics[width=\textwidth]{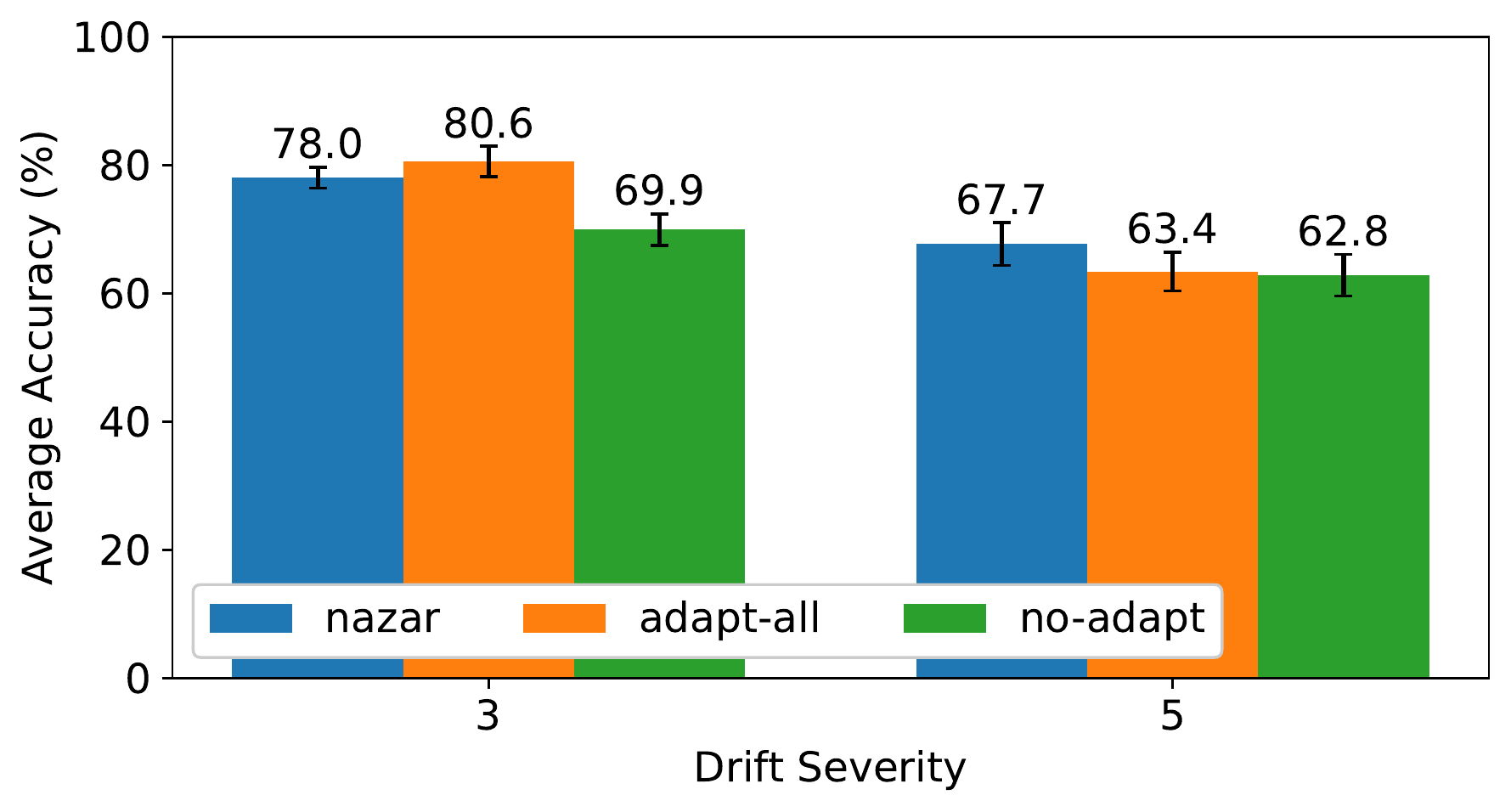}
   \subcaption{Average accuracy ($\alpha$ = 1).}
  \label{fig:imagenet_skewness}
  \end{subfigure}
\begin{subfigure}[t]{0.243\textwidth}
   \includegraphics[width=\textwidth]{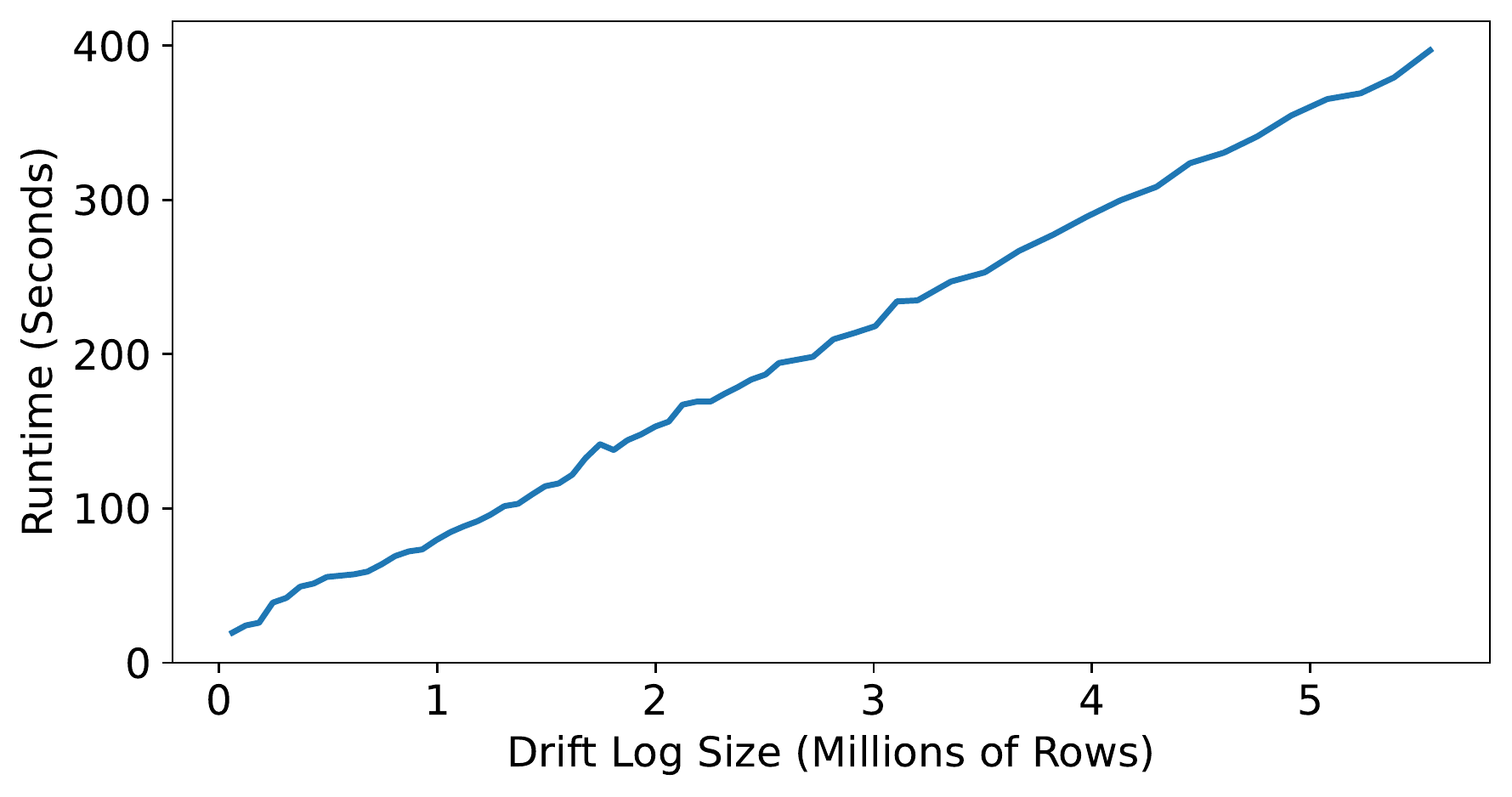} 
   \caption{Root cause analysis scalability.}
  \label{fig:scalability}
  \end{subfigure}
  \caption{Results of experiments on Animals dataset. ~\ref{fig:avg_acc_s3} and \ref{fig:avg_acc_s5} show the average accuracy with different drift severities under three strategies. \ref{fig:imagenet_skewness} shows the average accuracy with class skew of $\alpha$=1. ~\ref{fig:scalability} shows the scalability of \name's root analysis function.}
  \vspace{-1em}
  \label{fig:end-2-end_ablation}
\end{figure*}

\subsection{Evolving Drift Detection (Q4)}
\label{sec:interaction-eval}
We test how the by-cause adaptation influences the drift detector. Using the same experimental setup as the previous subsection, Figure~\ref{fig:detection-rate-before-after} displays the detection rate for each one of the drift types before adaptation and after adaptation. The after adaptation detection rate is measured for the adapted model that matches the drift type of the input image. The results show two interesting insights. First, before adaptation the detector is rather accurate in catching most of the sources of drift, but has some noise. Second, when it uses the appropriately adapted model, it is less likely to detect the data as drift, and exhibits the same detection probability for drifted data and clean data. 
Therefore, \name is not expected to frequently detect root causes again once it has adapted to them.

\subsection{End-to-End Workloads (Q5)}
\label{sec:end-to-end}
We now evaluate all components of \name end-to-end on our streaming workloads. We first evaluate the general performance on the cityscapes dataset and then test them on the more synthetic animal dataset under more severe drift conditions, \ie higher severity for weather drifts and class skew. 

\paragraph{Average accuracy.} We measure the average accuracy on all data averaged across the last 7 time windows on the cityscapes dataset in Figure~\ref{fig:cityscapes_8_avg}. For all model architectures, \name yields the highest average accuracies and the smallest standard deviations. The improvements in average accuracy compared with adapt-all are very significant: 10.1--19.4\%. 

We also measure the average accuracy of only the drifted data averaged across three types of data drift in Figure~\ref{fig:cityscapes_8_corr}, which shows even more significant accuracy improvements compared to adapt-all (49.5\% on ResNet18 and 37.6\% on ResNet34), which is because smaller models have a weaker ability to generalize on a mixed source of distributions. 

Figure~\ref{fig:cityscapes_8_trace} shows the cumulative average accuracy over time for all data and only for drifted images. The cumulative accuracy of \name continuously improves over time with each adaptation, while the one for adapt-all increases for one adaptation window, as the drift level is mild even in the no-adapt case, and then decreases in the following 3 adaptation windows, after which they slightly increase. This demonstrates the utility of adapt by-cause, which provides much more stable accuracy over time, both for clean and drifted data.

\paragraph{Benefit of set reduction and counterfactual analysis.} To find out if the set reduction and counterfactual analysis are effective in the end-to-end setting, we only run the FIM algorithm for root cause analysis, and find the average accuracy drops 1.3--9.7\%. 
Meanwhile, the number of stored BN versions (recall we only adapt the BN layer, so each BN layer represents a different adapted model) is much higher than \name, due to the redundant root causes. We show an example of the numbers of BN versions, from ResNet18, stored on user devices during each time window for FIM and \name in Figure~\ref{fig:bn_verions}, where the number of BN versions with \name is steady at 3 from the second time window. Note in this experiment we do not cap the number of model versions for \name. 

\paragraph{Adaptation frequency.}
We evaluate if the number of adaptation cycles affects adaptation performance.
We ran the end-to-end workloads by splitting them into 4 intervals instead of 8, and found the results stayed very consistent. 
The average accuracy across the three models improved by 1.2-3.8\%. 
\vspace{-0.5em}
\paragraph{Drift severity.}
We vary the severity of the corruptions on the animal dataset (Figure~\ref{fig:avg_acc_s3} and~\ref{fig:avg_acc_s5}), where a higher severity represents a higher level of drift, and severity is denoted by ``S''. When severity is higher, all three methods degrade on both accuracy metrics but \name still outperforms among them. \name's improvements compared to adapt-all is larger when the severity is higher (3.8--10.4\% better). 

\paragraph{Class skew.}
We evaluate the effect of class skew on \name with ResNet50, by setting the Zipfian parameter $\alpha=1$. 
The results are presented in Figure~\ref{fig:imagenet_skewness}. \name fails to outperform adapt-all on average accuracy when severity is set to 3 on the animal dataset, because \name has a narrower view of diverse image features than adapt-all, when the variability of classes in each location is largely constrained by class skew, and thus tends to overfit the model during adaptation. When we increase the variety of images during each adaptation by splitting the workload into 4 intervals instead of 8, \name outperforms both baselines and improves the average accuracy of 0.9\% compared with adapt-all. In addition, when the severity is higher, \name yields better average accuracy even when the number of time windows is 8, also shown in Figure~\ref{fig:imagenet_skewness} which offsets the impact from class skew.

\subsection{Runtime and Scalability (Q6)}
\label{sec:implementation-eval}
\paragraph{Runtime.}
Once a sufficient number of entries contain \OOD, \name should be able to generate adaptations in a reasonably timely manner to respond to the drift. We measure the latency of \name from the invocation of the root cause analysis function, through the model adaptations and up until the adapted models are written S3, and repeat the experiment four times. 
The average end-to-end latency is 49.5 minutes. The average root cause analysis runtime is only 46 seconds, while the rest of the time is consumed by the model adaptation. Note that this time can be shortened by adding additional or larger adaptation instances. 

\paragraph{Scalability.} 
\name's three components are designed to be highly scalable. The drift log relies on a scalable database (Aurora), and model adaptation can be easily parallelized on a large number of instances. Thus, the only component that could potentially become a scalability bottleneck is the root cause analysis function, since a single function needs to operate on a relatively large number of Aurora rows via a series of SQL queries.
Figure~\ref{fig:scalability} presents the runtime of the root cause analysis running on a single instance, as a function of the drift log size. 
We can see the relationship between the runtime and the number of rows in the drift log is completely linear. The reason is that the FIM algorithm is linear~\cite{apriori} and the number of candidate sets of attributes is greatly reduced for counterfactual analysis after the set reduction. 


%% file: datasets.tex
Our evaluation uses two configurable datasets: cityscapes and animal wildlife, which represent the datasets of two typical applications of \name.
Both datasets are emulated over a period between January 1st 2020 and April 21st, 2020, following a Kaggle dataset of historical weather~\cite{weather_kaggle}, which exhibits dynamic weather patterns. We apply synthetic weather-based data drifts (rain, snow and fog) based on the historic weather for different locations using data from a weather website~\cite{wunderground}. 
By default, we evenly divide the time for both datasets by 8, and run \name after each interval.
The drifts are applied as different types of random corruptions to the images. 
We parameterize the severity of different data drifts and the probability of an image experiencing them under the corresponding weather condition. By default, we set this probability to 1. 
For example, if an image from Hamburg is tagged to have been generated on January 18th, 2020, and the historical weather for that day indicated there was rain at some point during that day, by default, we apply a drift function for rain (\eg see Figure~\ref{fig:weather_comparison}) on that image.

\paragraph{Cityscapes.}
A potential application of \name is in driver automation and self-driving car models, which conduct on-device object classification and image segmentation \cite{DBLP:journals/corr/BojarskiTDFFGJM16}. Similar to prior work (Ekya~\cite{ekya}), we prepare an object detection dataset for a self-driving car application, built on top of the cityscapes dataset, which contains photos of traffic-related objects collected from driving cars from 50 cities in Europe~\cite{cityscapes}. Following Ekya's methodology, we pre-process cityscapes for object classification, resulting in 27,604 images, and use 14\% of the dataset for the initial training, 6\% for validation and 80\% for streaming new inferences. 
Since photos in each city are associated with sequence numbers, we assume users submit them based on their original order.
We assume the dataset at each location starts at January 1st, 2020 and ends at April 21st, 2020, and that the images are submitted for inference in equal intervals across these dates. 

\paragraph{Animals.}
In addition to cityscapes, which is a temporally-tagged dataset, we also generate a parameterized synthetic dataset that would allow us to methodologically configure our system.
For this dataset we are inspired by iNaturist~\cite{inaturalist}, an application for nature lovers to tag animals in the wild, which uses object classification to help users identify animal species in seconds. 
We emulate a geodistributed deployment of iNaturalist, where subscribers from different locations are taking photos and attempting to identify different animal species in the wild. We build this dataset on top of a subset of the classes from ImageNet~\cite{imagenet-localization}, which contains 201 classes of animals excluding pets. Each one of these classes contains on average 793 images for model training, 50 images for validation, and 500 images for streaming. We did not use the established iNaturist~\cite{46824} because both the distribution shift functions and adaptation methods were designed and tested for ImageNet~\cite{ImageNet-C,DBLP:journals/corr/abs-2006-10726,DBLP:journals/corr/abs-2110-09506} and we want to conduct a fair comparison between \name and prior adaptation approaches. 

We emulate 7 locations from different continents: New York, Tibet, Beijing, New South Wales, United Kingdom and Quebec, where each location contains a different distribution of animal species. Each location contains a configurable number of user devices with a configurable arrival pattern of inference requests. By default, we set the number of devices to 16 for each location and the request pattern to be a Poisson distribution with a mean of two images per day per device. 

\paragraph{Class skew.} We introduce another source of \OOD, \emph{class skew}. We observe that a model's accuracy of images from different classes varies (Figure~\ref{fig:sorted_201cls}), even though the number of training images for each class is similar. Therefore, if a certain location exhibits a higher proportion of images from lower-accuracy classes, the model's accuracy may suffer. To emulate this effect, we generate the distribution of classes for each location by randomly assigning probabilities for each class under the Zipf distribution, where the higher the Zipf parameter $\alpha$ is, the more skewed the class distribution. 
By default, we set $\alpha$ to be 0 (uniform) but also evaluate under a high skew ($\alpha=1$). 

%% file: conclusions.tex
\section{Conclusions}

\name is the first end-to-end monitoring and adaptation system for mobile-based ML deployments. We conclude that chaining together drift detection run on the device, with root cause analysis and by-cause adaptation run in the cloud, provides high accuracy, even without user input. Interesting avenues for future work are adapting \name to distributed federated learning, and developing techniques for better protecting user privacy. 

%% file: usenix.bbl
\begin{thebibliography}{10}

\bibitem{awsec2p3}
Amazon {EC2 P3} instances.
\newblock \url{https://aws.amazon.com/ec2/instance-types/p3/}.
\newblock Accessed: April 27, 2023.

\bibitem{awsrds}
Amazon {RDS} instance types.
\newblock \url{https://aws.amazon.com/rds/instance-types/}.
\newblock Accessed: April 27, 2023.

\bibitem{awslambda}
{AWS Lambda}.
\newblock \url{https://aws.amazon.com/lambda/}.
\newblock Accessed: April 27, 2023.

\bibitem{imagenet-localization}
{ImageNet Object Localization Challenge}.
\newblock
  \url{https://www.kaggle.com/c/imagenet-object-localization-challenge}, 2017.

\bibitem{weather_kaggle}
Historical daily weather data 2020.
\newblock
  \url{https://www.kaggle.com/datasets/vishalvjoseph/weather-dataset-for-covid19-predictions},
  2020.
\newblock Accessed: Apr 12, 2021.

\bibitem{inaturalist}
{iNaturalist}.
\newblock {https://www.inaturalist.org/}, 2023.

\bibitem{torchmobile}
Pytorch android examples.
\newblock \url{https://github.com/pytorch/android-demo-app}, 2023.
\newblock Accessed: May 3, 2023.

\bibitem{tflite}
{TensorFlow Lite}.
\newblock \url{https://www.tensorflow.org/lite}, 2023.

\bibitem{wunderground}
Weather underground.
\newblock \url{https://www.wunderground.com/}, 2023.
\newblock Accessed: Mar 8, 2023.

\bibitem{diff}
{\sc Abuzaid, F., Kraft, P., Suri, S., Gan, E., Xu, E., Shenoy, A.,
  Ananthanarayan, A., Sheu, J., Meijer, E., Wu, X., Naughton, J., Bailis, P.,
  and Zaharia, M.}
\newblock {DIFF}: A relational interface for large-scale data explanation.
\newblock {\em Proc. VLDB Endow. 12}, 4 (dec 2018), 419–432.

\bibitem{item-sets}
{\sc Agrawal, R., Imieli\'{n}ski, T., and Swami, A.}
\newblock Mining association rules between sets of items in large databases.
\newblock In {\em Proceedings of the 1993 ACM SIGMOD International Conference
  on Management of Data\/} (New York, NY, USA, 1993), SIGMOD '93, Association
  for Computing Machinery, p.~207–216.

\bibitem{agrawal1996fast}
{\sc Agrawal, R., Mannila, H., Srikant, R., Toivonen, H., Verkamo, A.~I.,
  et~al.}
\newblock Fast discovery of association rules.
\newblock {\em Advances in knowledge discovery and data mining 12}, 1 (1996),
  307--328.

\bibitem{apriori}
{\sc Agrawal, R., Srikant, R., et~al.}
\newblock Fast algorithms for mining association rules.
\newblock In {\em Proc. 20th int. conf. very large data bases, VLDB\/} (1994),
  vol.~1215, Santiago, Chile, pp.~487--499.

\bibitem{ekya}
{\sc Bhardwaj, R., Xia, Z., Ananthanarayanan, G., Jiang, J., Karianakis, N.,
  Shu, Y., Hsieh, K., Bahl, V., and Stoica, I.}
\newblock Ekya: Continuous learning of video analytics models on edge compute
  servers.
\newblock {\em CoRR abs/2012.10557\/} (2020).

\bibitem{DBLP:journals/corr/BojarskiTDFFGJM16}
{\sc Bojarski, M., Testa, D.~D., Dworakowski, D., Firner, B., Flepp, B., Goyal,
  P., Jackel, L.~D., Monfort, M., Muller, U., Zhang, J., Zhang, X., Zhao, J.,
  and Zieba, K.}
\newblock End to end learning for self-driving cars.
\newblock {\em CoRR abs/1604.07316\/} (2016).

\bibitem{borgelt2003efficient}
{\sc Borgelt, C.}
\newblock Efficient implementations of apriori and eclat.
\newblock In {\em FIMI’03: Proceedings of the IEEE ICDM workshop on frequent
  itemset mining implementations\/} (2003), Citeseer, p.~90.

\bibitem{borgelt2005implementation}
{\sc Borgelt, C.}
\newblock An implementation of the fp-growth algorithm.
\newblock In {\em Proceedings of the 1st international workshop on open source
  data mining: frequent pattern mining implementations\/} (2005), pp.~1--5.

\bibitem{chen2019federated}
{\sc Chen, M., Mathews, R., Ouyang, T., and Beaufays, F.}
\newblock Federated learning of out-of-vocabulary words.
\newblock {\em arXiv preprint arXiv:1903.10635\/} (2019).

\bibitem{cidon2021characterizing}
{\sc Cidon, E., Pergament, E., Asgar, Z., Cidon, A., and Katti, S.}
\newblock Characterizing and taming model instability across edge devices.
\newblock {\em Proceedings of Machine Learning and Systems 3\/} (2021).

\bibitem{cityscapes}
{\sc Cordts, M., Omran, M., Ramos, S., Rehfeld, T., Enzweiler, M., Benenson,
  R., Franke, U., Roth, S., and Schiele, B.}
\newblock The cityscapes dataset for semantic urban scene understanding.
\newblock {\em CoRR abs/1604.01685\/} (2016).

\bibitem{FMI}
{\sc Fowlkes, E.~B., and Mallows, C.~L.}
\newblock A method for comparing two hierarchical clusterings.
\newblock {\em Journal of the American statistical association 78}, 383 (1983),
  553--569.

\bibitem{french1999catastrophic}
{\sc French, R.~M.}
\newblock Catastrophic forgetting in connectionist networks.
\newblock {\em Trends in cognitive sciences 3}, 4 (1999), 128--135.

\bibitem{ghosh}
{\sc Ghosh, A., Chung, J., Yin, D., and Ramchandran, K.}
\newblock An efficient framework for clustered federated learning.
\newblock {\em Advances in Neural Information Processing Systems 33\/} (2020),
  19586--19597.

\bibitem{mistify}
{\sc Guo, P., Hu, B., and Hu, W.}
\newblock Mistify: Automating {DNN} model porting for on-device inference at
  the edge.
\newblock In {\em NSDI\/} (2021), pp.~705--719.

\bibitem{han2000mining}
{\sc Han, J., Pei, J., and Yin, Y.}
\newblock Mining frequent patterns without candidate generation.
\newblock {\em ACM sigmod record 29}, 2 (2000), 1--12.

\bibitem{hao2022tale}
{\sc Hao, W., Awatramani, A., Hu, J., Mao, C., Chen, P.-C., Cidon, E., Cidon,
  A., and Yang, J.}
\newblock A tale of two models: Constructing evasive attacks on edge models.
\newblock {\em Proceedings of Machine Learning and Systems 4\/} (2022),
  414--429.

\bibitem{fedml}
{\sc He, C., Li, S., So, J., Zeng, X., Zhang, M., Wang, H., Wang, X.,
  Vepakomma, P., Singh, A., Qiu, H., et~al.}
\newblock {FedML}: A research library and benchmark for federated machine
  learning.
\newblock {\em arXiv preprint arXiv:2007.13518\/} (2020).

\bibitem{DBLP:journals/corr/abs-1812-05720}
{\sc Hein, M., Andriushchenko, M., and Bitterwolf, J.}
\newblock Why relu networks yield high-confidence predictions far away from the
  training data and how to mitigate the problem.
\newblock {\em CoRR abs/1812.05720\/} (2018).

\bibitem{ImageNet-C}
{\sc Hendrycks, D., and Dietterich, T.~G.}
\newblock Benchmarking neural network robustness to common corruptions and
  perturbations.
\newblock {\em CoRR abs/1903.12261\/} (2019).

\bibitem{MSP}
{\sc Hendrycks, D., and Gimpel, K.}
\newblock A baseline for detecting misclassified and out-of-distribution
  examples in neural networks.
\newblock {\em CoRR abs/1610.02136\/} (2016).

\bibitem{OE}
{\sc Hendrycks, D., Mazeika, M., and Dietterich, T.~G.}
\newblock Deep anomaly detection with outlier exposure.
\newblock {\em CoRR abs/1812.04606\/} (2018).

\bibitem{ssl-ood}
{\sc Hendrycks, D., Mazeika, M., Kadavath, S., and Song, D.}
\newblock Using self-supervised learning can improve model robustness and
  uncertainty.
\newblock {\em CoRR abs/1906.12340\/} (2019).

\bibitem{46824}
{\sc Horn, G.~V., Aodha, O.~M., Song, Y., Cui, Y., Sun, C., Shepard, A., Adam,
  H., Perona, P., and Belongie, S.}
\newblock The inaturalist species classification and detection dataset.
\newblock In {\em CVPR\/} (2018).

\bibitem{Godin}
{\sc Hsu, Y., Shen, Y., Jin, H., and Kira, Z.}
\newblock Generalized {ODIN:} detecting out-of-distribution image without
  learning from out-of-distribution data.
\newblock {\em CoRR abs/2002.11297\/} (2020).

\bibitem{BatchNorm}
{\sc Ioffe, S., and Szegedy, C.}
\newblock Batch normalization: Accelerating deep network training by reducing
  internal covariate shift.
\newblock {\em CoRR abs/1502.03167\/} (2015).

\bibitem{fedscale}
{\sc Lai, F., Dai, Y., Singapuram, S., Liu, J., Zhu, X., Madhyastha, H., and
  Chowdhury, M.}
\newblock Fedscale: Benchmarking model and system performance of federated
  learning at scale.
\newblock In {\em International Conference on Machine Learning\/} (2022), PMLR,
  pp.~11814--11827.

\bibitem{oort}
{\sc Lai, F., Zhu, X., Madhyastha, H.~V., and Chowdhury, M.}
\newblock Oort: Efficient federated learning via guided participant selection.
\newblock In {\em 15th {USENIX} Symposium on Operating Systems Design and
  Implementation ({OSDI} 21)\/} (July 2021), {USENIX} Association, pp.~19--35.

\bibitem{lee2017training}
{\sc Lee, K., Lee, H., Lee, K., and Shin, J.}
\newblock Training confidence-calibrated classifiers for detecting
  out-of-distribution samples.
\newblock {\em arXiv preprint arXiv:1711.09325\/} (2017).

\bibitem{MD}
{\sc Lee, K., Lee, K., Lee, H., and Shin, J.}
\newblock A simple unified framework for detecting out-of-distribution samples
  and adversarial attacks.
\newblock In {\em NeurIPS\/} (2018).

\bibitem{lewis1973counterfactuals}
{\sc Lewis, D.}
\newblock {\em Counterfactuals}.
\newblock John Wiley \& Sons, 2013.

\bibitem{li2017learning}
{\sc Li, Z., and Hoiem, D.}
\newblock Learning without forgetting.
\newblock {\em IEEE transactions on pattern analysis and machine intelligence
  40}, 12 (2017), 2935--2947.

\bibitem{odin}
{\sc Liang, S., Li, Y., and Srikant, R.}
\newblock Principled detection of out-of-distribution examples in neural
  networks.
\newblock {\em CoRR abs/1706.02690\/} (2017).

\bibitem{sagemaker}
{\sc Liberty, E., Karnin, Z., Xiang, B., Rouesnel, L., Coskun, B., Nallapati,
  R., Delgado, J., Sadoughi, A., Astashonok, Y., Das, P., et~al.}
\newblock Elastic machine learning algorithms in amazon sagemaker.
\newblock In {\em Proceedings of the 2020 ACM SIGMOD International Conference
  on Management of Data\/} (2020), pp.~731--737.

\bibitem{energy-ood}
{\sc Liu, W., Wang, X., Owens, J.~D., and Li, Y.}
\newblock Energy-based out-of-distribution detection.
\newblock {\em CoRR abs/2010.03759\/} (2020).

\bibitem{maltoni2019continuous}
{\sc Maltoni, D., and Lomonaco, V.}
\newblock Continuous learning in single-incremental-task scenarios.
\newblock {\em Neural Networks 116\/} (2019), 56--73.

\bibitem{mccloskey1989catastrophic}
{\sc McCloskey, M., and Cohen, N.~J.}
\newblock Catastrophic interference in connectionist networks: The sequential
  learning problem.
\newblock In {\em Psychology of learning and motivation}, vol.~24. Elsevier,
  1989, pp.~109--165.

\bibitem{TFX}
{\sc Modi, A.~N., Koo, C.~Y., Foo, C.~Y., Mewald, C., Baylor, D.~M., Breck, E.,
  Cheng, H.-T., Wilkiewicz, J., Koc, L., Lew, L., Zinkevich, M.~A., Wicke, M.,
  Ispir, M., Polyzotis, N., Fiedel, N., Haykal, S.~E., Whang, S., Roy, S.,
  Ramesh, S., Jain, V., Zhang, X., and Haque, Z.}
\newblock Tfx: A tensorflow-based production-scale machine learning platform.
\newblock In {\em KDD 2017\/} (2017).

\bibitem{DBLP:journals/corr/NguyenYC14}
{\sc Nguyen, A.~M., Yosinski, J., and Clune, J.}
\newblock Deep neural networks are easily fooled: High confidence predictions
  for unrecognizable images.
\newblock {\em CoRR abs/1412.1897\/} (2014).

\bibitem{parisi2019continual}
{\sc Parisi, G.~I., Kemker, R., Part, J.~L., Kanan, C., and Wermter, S.}
\newblock Continual lifelong learning with neural networks: A review.
\newblock {\em Neural networks 113\/} (2019), 54--71.

\bibitem{paulik2021federated}
{\sc Paulik, M., Seigel, M., Mason, H., Telaar, D., Kluivers, J., van Dalen,
  R., Lau, C.~W., Carlson, L., Granqvist, F., Vandevelde, C., et~al.}
\newblock Federated evaluation and tuning for on-device personalization: System
  design \& applications.
\newblock {\em arXiv preprint arXiv:2102.08503\/} (2021).

\bibitem{pei2000closet}
{\sc Pei, J., Han, J., Mao, R., et~al.}
\newblock Closet: An efficient algorithm for mining frequent closed itemsets.
\newblock In {\em ACM SIGMOD workshop on research issues in data mining and
  knowledge discovery\/} (2000), vol.~4, pp.~21--30.

\bibitem{914830}
{\sc Pei, J., Han, J., Mortazavi-Asl, B., Pinto, H., Chen, Q., Dayal, U., and
  Hsu, M.-C.}
\newblock Prefixspan,: mining sequential patterns efficiently by
  prefix-projected pattern growth.
\newblock In {\em Proceedings 17th International Conference on Data
  Engineering\/} (2001), pp.~215--224.

\bibitem{qiu2022ml}
{\sc Qiu, H., Vavelidou, I., Li, J., Pergament, E., Warden, P., Chinchali, S.,
  Asgar, Z., and Katti, S.}
\newblock {ML-EXray}: Visibility into {ML} deployment on the edge.
\newblock {\em Proceedings of Machine Learning and Systems 4\/} (2022),
  337--351.

\bibitem{rabanser2019failing}
{\sc Rabanser, S., G{\"u}nnemann, S., and Lipton, Z.}
\newblock Failing loudly: An empirical study of methods for detecting dataset
  shift.
\newblock {\em Advances in Neural Information Processing Systems 32\/} (2019).

\bibitem{ramaswamy2019federated}
{\sc Ramaswamy, S., Mathews, R., Rao, K., and Beaufays, F.}
\newblock Federated learning for emoji prediction in a mobile keyboard.
\newblock {\em arXiv preprint arXiv:1906.04329\/} (2019).

\bibitem{rebuffi2017icarl}
{\sc Rebuffi, S.-A., Kolesnikov, A., Sperl, G., and Lampert, C.~H.}
\newblock icarl: Incremental classifier and representation learning.
\newblock In {\em Proceedings of the IEEE conference on Computer Vision and
  Pattern Recognition\/} (2017), pp.~2001--2010.

\bibitem{sculley2010web}
{\sc Sculley, D.}
\newblock Web-scale {K}-means clustering.
\newblock In {\em Proceedings of the 19th international conference on World
  wide web\/} (2010), pp.~1177--1178.

\bibitem{shankar2023moving}
{\sc Shankar, S., Fawaz, L., Gyllstrom, K., and Parameswaran, A.~G.}
\newblock Moving fast with broken data.
\newblock {\em arXiv preprint arXiv:2303.06094\/} (2023).

\bibitem{shankar2022operationalizing}
{\sc Shankar, S., Garcia, R., Hellerstein, J.~M., and Parameswaran, A.~G.}
\newblock Operationalizing machine learning: An interview study, 2022.

\bibitem{shankar2022rethinking}
{\sc Shankar, S., Herman, B., and Parameswaran, A.~G.}
\newblock Rethinking streaming machine learning evaluation.
\newblock {\em arXiv preprint arXiv:2205.11473\/} (2022).

\bibitem{DBLP:journals/corr/abs-2108-13557}
{\sc Shankar, S., and Parameswaran, A.~G.}
\newblock Towards observability for machine learning pipelines.
\newblock {\em CoRR abs/2108.13557\/} (2021).

\bibitem{shmelkov2017incremental}
{\sc Shmelkov, K., Schmid, C., and Alahari, K.}
\newblock Incremental learning of object detectors without catastrophic
  forgetting.
\newblock In {\em Proceedings of the IEEE international conference on computer
  vision\/} (2017), pp.~3400--3409.

\bibitem{CSI}
{\sc Tack, J., Mo, S., Jeong, J., and Shin, J.}
\newblock {CSI:} novelty detection via contrastive learning on distributionally
  shifted instances.
\newblock {\em CoRR abs/2007.08176\/} (2020).

\bibitem{alibi-detect}
{\sc Van~Looveren, A., Klaise, J., Vacanti, G., Cobb, O., Scillitoe, A., and
  Samoilescu, R.}
\newblock Alibi detect: Algorithms for outlier, adversarial and drift
  detection, 2019.

\bibitem{aurora}
{\sc Verbitski, A., Gupta, A., Saha, D., Brahmadesam, M., Gupta, K., Mittal,
  R., Krishnamurthy, S., Maurice, S., Kharatishvili, T., and Bao, X.}
\newblock Amazon {Aurora}: Design considerations for high throughput
  cloud-native relational databases.
\newblock In {\em Proceedings of the 2017 ACM International Conference on
  Management of Data\/} (2017), pp.~1041--1052.

\bibitem{DBLP:journals/corr/abs-2006-10726}
{\sc Wang, D., Shelhamer, E., Liu, S., Olshausen, B.~A., and Darrell, T.}
\newblock Fully test-time adaptation by entropy minimization.
\newblock {\em CoRR abs/2006.10726\/} (2020).

\bibitem{wang2023flint}
{\sc Wang, E., Kannan, A., Liang, Y., Chen, B., and Chowdhury, M.}
\newblock {FLINT}: A platform for federated learning integration.
\newblock {\em arXiv preprint arXiv:2302.12862\/} (2023).

\bibitem{widmer1996learning}
{\sc Widmer, G., and Kubat, M.}
\newblock Learning in the presence of concept drift and hidden contexts.
\newblock {\em Machine learning 23\/} (1996), 69--101.

\bibitem{wu2019machine}
{\sc Wu, C.-J., Brooks, D., Chen, K., Chen, D., Choudhury, S., Dukhan, M.,
  Hazelwood, K., Isaac, E., Jia, Y., Jia, B., et~al.}
\newblock Machine learning at {Facebook}: Understanding inference at the edge.
\newblock In {\em 2019 IEEE International Symposium on High Performance
  Computer Architecture (HPCA)\/} (2019), IEEE, pp.~331--344.

\bibitem{yin2018feature}
{\sc Yin, X., Yu, X., Sohn, K., Liu, X., and Chandraker, M.}
\newblock Feature transfer learning for deep face recognition with
  under-represented data.
\newblock {\em arXiv preprint arXiv:1803.09014\/} (2018).

\bibitem{DBLP:journals/corr/abs-2110-09506}
{\sc Zhang, M., Levine, S., and Finn, C.}
\newblock {MEMO:} test time robustness via adaptation and augmentation.
\newblock {\em CoRR abs/2110.09506\/} (2021).

\end{thebibliography}
